\documentclass[10pt,journal,compsoc]{IEEEtran}

\usepackage{graphicx}
\usepackage{float}
\usepackage[caption=false]{subfig}
\usepackage[justification=centering]{caption}
\usepackage{verbatim}
\usepackage{algorithmic}
\usepackage{algorithm}
\usepackage{multirow}
\usepackage{amsmath,amssymb}
\usepackage{color}
\usepackage{ulem}
\usepackage{url}
\usepackage{hyperref}

\iftrue
\usepackage{xargs}
\usepackage[colorinlistoftodos,prependcaption,textsize=tiny]{todonotes}
\newenvironment{newljc}[1]{\color{#1}}{\ignorespacesafterend}
\newcommand{\nnewljc}[1]{\textcolor{black}{#1}}

\newif\ifshowcomments
\showcommentsfalse 

\ifshowcomments
  \newcommand{\revbegin}[2]{\textcolor{red}{[Reviewer#1-#2:Begin]}}
  \newcommand{\revend}[2]{\textcolor{blue}{[Reviewer#1-#2:End]}}
\else
  \newcommand{\revbegin}[2]{}
  \newcommand{\revend}[2]{}
\fi

\newcommandx{\info}[2][1=]{\todo[linecolor=OliveGreen,backgroundcolor=OliveGreen!25,bordercolor=OliveGreen,#1]{#2}}
\newcommand{\delline}[1]{\textcolor{blue}{\sout{#1}}}
\newcommand{\del}[1]{\delline{#1}}
\else
\newenvironment{newljc}[1]{#1}{\ignorespacesafterend}
\newcommand{\nnewljc}[1]{#1}

\newcommand{\revbegin}[2]{}
\newcommand{\revend}[2]{}

\newcommand{\info}[1]{}
\newcommand{\delline}[1]{}
\newcommand{\del}[1]{}
\fi

\IEEEpubid{\begin{minipage}{\textwidth}\ 
         \copyright 2025 IEEE. Personal use of this material is permitted.  Permission from IEEE must be obtained for all other uses, in any current or future media, including reprinting/republishing this material for advertising or promotional purposes, creating new collective works, for resale or redistribution to servers or lists, or reuse of any copyrighted component of this work in other works.
\end{minipage}}

\begin{document}

\title{We Can Always Catch You:  \\ Detecting Adversarial Patched Objects WITH or WITHOUT Signature}

\author{Jiachun Li,
        Jianan Feng,
        Jianjun Huang,   
        and Bin Liang
\thanks{This work is supported in part by National Natural Science Foundation of China (NSFC) 
under grants 62272465 and 62272464.} 
\thanks{J. Li, J. Feng, J. Huang and B. Liang, are with the School of Information, Renmin University of China, Beijing 100872, China; and also with Key Laboratory of DEKE (Renmin University of China), MOE, China. E-mail: \{jclee, jiananfeng, hjj, liangb\}@ruc.edu.cn.
}
\thanks{Bin Liang (liangb@ruc.edu.cn) is the corresponding author.}
}



\IEEEtitleabstractindextext{%
\begin{abstract}
Recently, object detection has proven vulnerable to adversarial patch attacks. The attackers holding a specially crafted patch can hide themselves from state-of-the-art detectors, e.g., YOLO, even in the physical world. This attack can bring serious security threats, such as escaping from surveillance cameras. How to effectively detect this kind of adversarial examples to catch potential attacks has become an important problem. In this paper, we propose two detection methods: the signature-based method and the signature-independent method. First, we identify two signatures of existing adversarial patches that can be utilized to precisely locate patches within adversarial examples. By employing the signatures, a fast signature-based method is developed to detect the adversarial objects. Second, we present a robust signature-independent method based on the \textit{content semantics consistency} of model outputs. Adversarial objects violate this consistency, appearing locally but disappearing globally, while benign ones remain consistently present. The experiments demonstrate that two proposed methods can effectively detect attacks both in the digital and physical world. These methods each offer distinct advantage. Specifically, the signature-based method is capable of real-time detection, while the signature-independent method can detect unknown adversarial patch attacks and makes defense-aware attacks almost impossible to perform.

\end{abstract}

\begin{IEEEkeywords}
Object detection, Adversarial patch attack, Detection method.
\end{IEEEkeywords}}

\maketitle

\IEEEdisplaynontitleabstractindextext
\IEEEpeerreviewmaketitle

\section{Introduction}\label{sec:introduction}
One of the most important applications of deep learning is object detection \cite{scaled-yolov4,YOLOR,2016Yolov1,yolo9000,2017FasterRCNN, yolov8}, which is designed to identify and locate the instances of specific target classes, e.g., humans and cars, in images or videos.
Unsurprisingly, object detection models have become the attack target in recent years. Some attack methods \cite{thys2019fooling,wu2020cloak,xu2020T-shirt} have been proposed to fool object detectors, making the target object evade detection. Some of them can even successfully launch evasion attacks in the physical world rather than only in the digital world.  
Thys et al. \cite{thys2019fooling} is the first to propose an adversarial patch generation algorithm to fool YOLO \cite{yolo9000} on person object. 
As they demonstrated, a person with the patch can hide from the surveillance camera equipped with a YOLO detector. The adversarial patch can also be directly printed on the T-shirts to fool the object detection models \cite{wu2020cloak,xu2020T-shirt}. We believe that more new attack methods will emerge in the future.

An important question arises naturally as \textit{how we can effectively detect the known and unknown adversarial patch attacks}. 

In this study, our first observation is that the adversarial patches generated by the existing technique \cite{thys2019fooling} possess distinct signatures, i.e. high region entropy and high Guided Grad-CAM score, which can be leveraged to effectively identify the patch pixels from the target video frames (which can be regarded as a sequence of images).

\IEEEpubidadjcol
Based on the observation, a straightforward but effective \textit{signature-based} adversarial object detection method is proposed. We utilize the signatures to identify the adversarial patch in the image and filter it out.
The object detection is performed on the produced image instead of the original one to suppress the effect of the adversarial patch as far as possible. The experiment demonstrates that our technique can effectively detect hidden objects in digital world and physical world, and it is capable of real-time detection.

\begin{figure}[!t]
    \centering
    \subfloat[][a benign example]{
        \includegraphics[width=3cm, height=4cm]{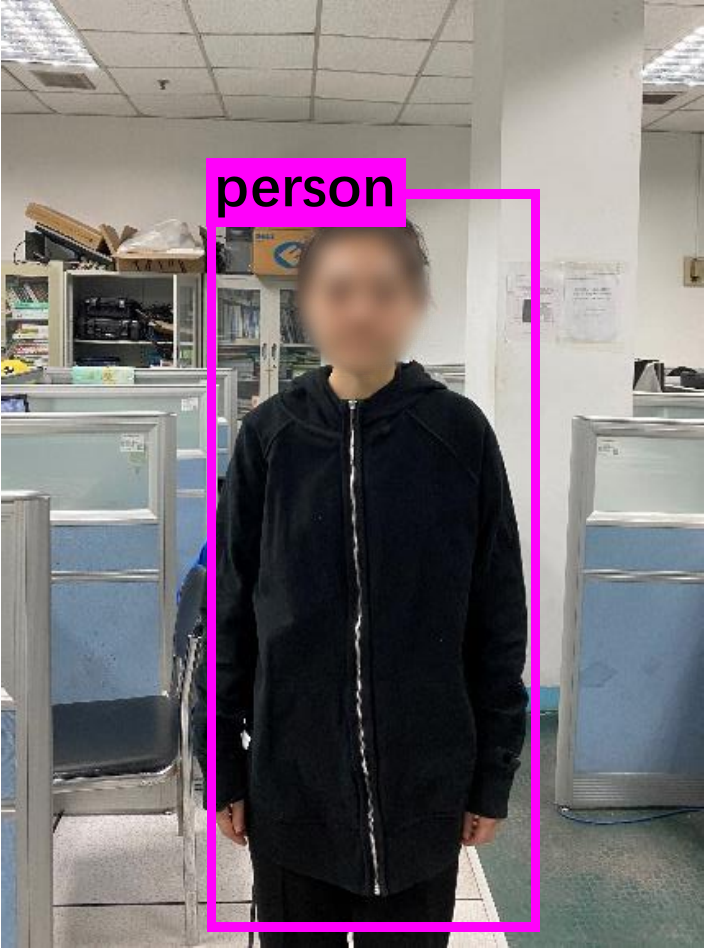}
        \label{fig:1a}
        }
    \subfloat[][an adversarial example generated with \cite{thys2019fooling}]{
        \includegraphics[width=3cm, height=4cm]{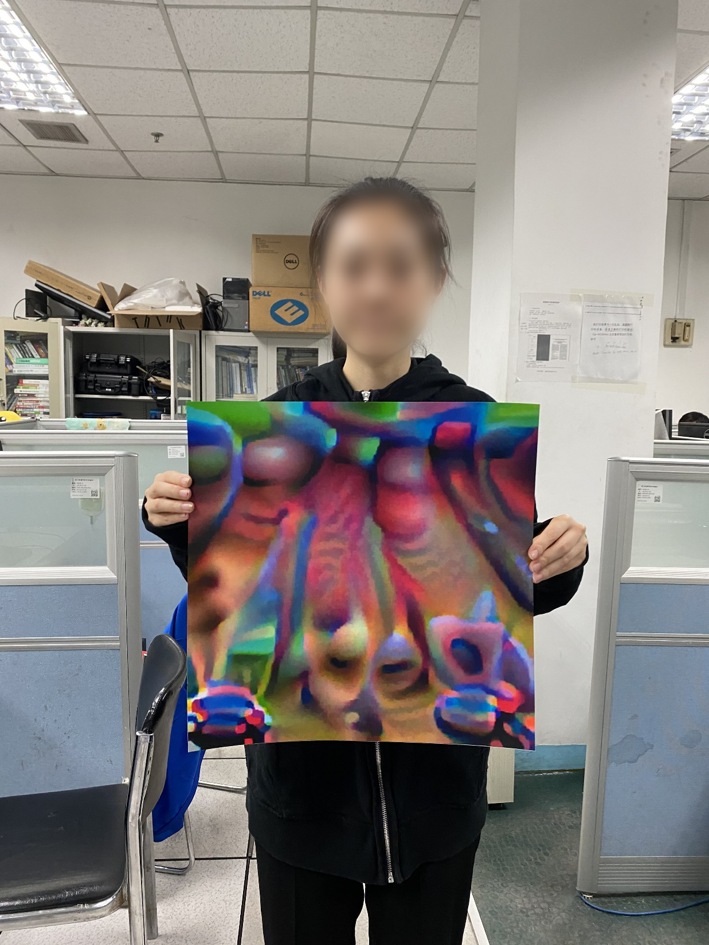}
        \label{fig:1b}
        }
    \vspace{-3mm} \\ 
    \subfloat[][detecting \subref{fig:1b} with the signature-based method]{
        \includegraphics[width=3cm, height=4cm]{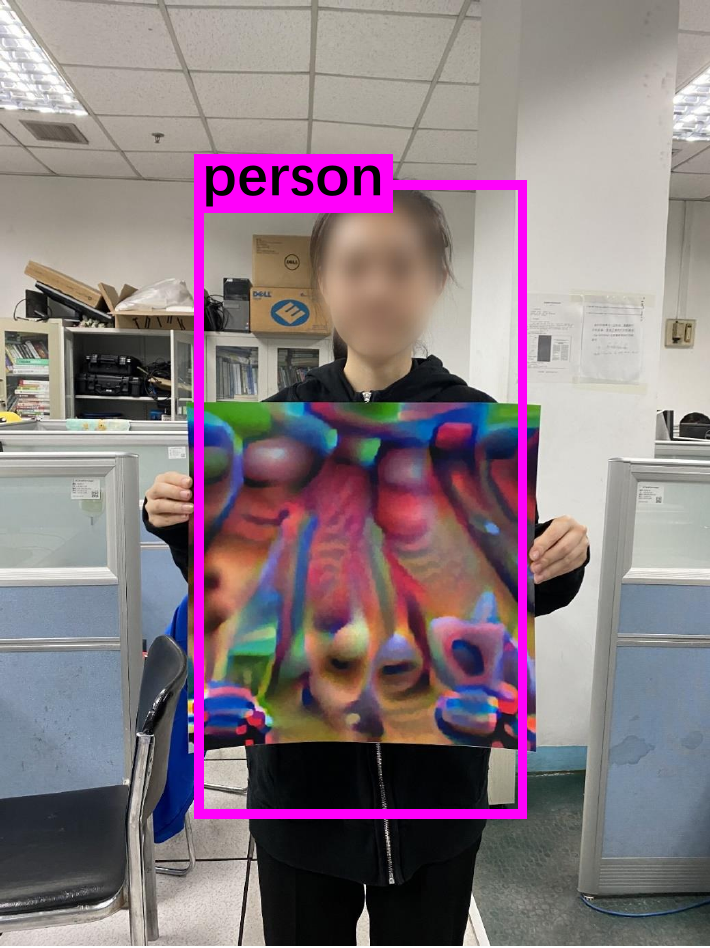}
        \label{fig:1c}
        }
    \subfloat[][detecting \subref{fig:1b} with the signature-independent method]{
        \includegraphics[width=3cm, height=4cm]{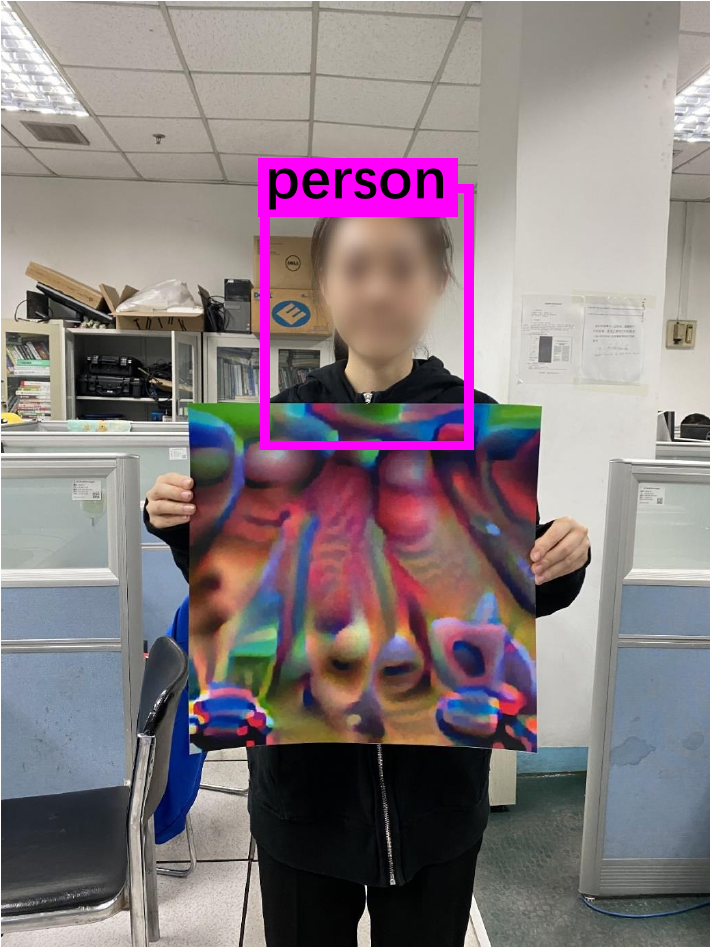}
        \label{fig:1e}
        }
    \caption{Attack and defense examples.}
\end{figure}

Taking YOLO as the target model, from Figure~\ref{fig:1a}, we can see that YOLO can successfully detect a person and mark her with a rectangular pink box. However, in Figure~\ref{fig:1b}, she can hide from the detector by holding an adversarial patch generated with the algorithm presented in \cite{thys2019fooling}. By integrating the signature-based defense technique, the YOLO model can detect the person from the filtered input, as shown in Figure~\ref{fig:1c}.

The signature-based method is fast and straightforward. However, it can only detect known attacks. There remains a possibility that it could be compromised by defense-aware attacks. Therefore, we propose the \textit{signature-independent} detection, which provides robustness against unknown attacks and inherently resists defense-aware attempts.
The method is based on an insightful observation that \textit{the local and global content semantics is inconsistent in adversarial patched examples}. In a benign example, if an object can be detected from a part of it, the object can also be detected from a larger part. However, in an adversarial example, the object may be invisible to the detector once the larger part includes enough adversarial pixels. In other words, the adversarial object can \textit{appear locally but disappear globally}. 
We design a region growing algorithm to check whether there is this kind of inconsistency in the current input and detect potential adversarial examples both in the digital world and the physical world. As shown in Figures~\ref{fig:1e}, the attacks in the physical world can be successfully detected. 
Our experiment also demonstrates the concept of content semantics consistency is general and can be adopted to detect other types of local perturbation attacks, e.g., \cite{brown2017advpatch}, \cite{xu2020T-shirt} and \cite{huang2020UPC}.

Note that the two proposed detection methods can be used for different scenarios. The signature-based one can be adopted in efficiency-sensitive scenarios. Due to involving semantics analysis, the signature-independent method is inevitably relatively slow. 
However, it can be used to catch unknown attacks via sampling or be implemented in parallel. When an unknown attack is caught, we can also upgrade the signature-based method via identifying new leverageable signatures. We believe that the two methods can be combined to provide a comprehensive protection against adversarial patch attacks. Furthermore, the two methods can also provide a concrete explanation about why the target objects can evade the detector, i.e., identifying the suspect pixels and regions that make the objects hidden.

In summary, the main contributions of this paper are as follows:

\begin{itemize}
\setlength{\itemsep}{0pt}
\setlength{\parsep}{0pt}
\setlength{\parskip}{0pt}
\item A fast and effective signature-based technique for detecting adversarial patches is proposed, which is time-efficient and can integrate seamlessly with existing object detection models.
\vspace{1mm}
\item A robust and general signature-independent technique for detecting adversarial patches is proposed based on the analysis of the local and global content semantics consistency. It is able to detect unknown adversarial patch attacks and is almost impossible to produce defense-aware attacks.
\vspace{1mm}
\item Our data set is available at:

\href{https://drive.google.com/drive/folders/10LE3DwuVNuBe5HrUtEjaz8GFY6vBLkNC?usp=drive\_link}{https://drive.google.com/drive/folders/10LE3Dwu
VNuBe5HrUtEjaz8GFY6vBLkNC?usp=drive\_link}. 

\end{itemize}

\section{Preliminary}
\label{sec:background}

\subsection{Background Knowledge} 

\subsubsection{Object detection} 
We take YOLO \cite{yolo9000}, one of the most popular object detection models, as an example. 
YOLO divides the input image into an $S \times S$ grid. Each grid cell predicts $B$ bounding boxes, and each bounding box is given five predictions: $x$, $y$, $w$, $h$ and \textit{object confidence}. The former four determine the coordinates and size of the box, and the object confidence reflects the probability that the box contains an object. Besides, for each cell, YOLO outputs $P$ conditional \textit{class probabilities}, the greatest one of which will be used to determine the class of the detected object. An object may cross multiple grid cells. However, only one cell is responsible for detecting the object, which contains the center of the target object.
At the test time, for each box, the product of the object confidence and the class probabilities are used to determine whether there is an object and which class the object belongs to. 

\subsubsection{Adversarial patch attacks} 
Thys et al. \cite{thys2019fooling} propose the concept of the adversarial patch attack on object detection. The adversarial patch is a small printable attack vector picture, which can be introduced into the input example like a ''patch''. They design an optimization process to generate the adversarial patch targeting YOLO, by reducing the object confidence of the bounding box.
The optimization goal consists of three components: the maximum object confidence $L_{obj}$, the non-printability score $L_{nps}$, and the total image variation $L_{tv}$. The total loss function of the optimization is the weighted sum of the three losses, as given in Equation~\ref{eq:Loss}.

\begin{equation}
    L = L_{obj} + \alpha L_{nps} + \beta L_{tv}
    \label{eq:Loss}
\end{equation}

\subsubsection{GrabCut} 
GrabCut \cite{GrabCut2004} is a traditional algorithm for image segmentation. A significant advantage of GrabCut \cite{GrabCut2004} is its ability to precisely segment adversarial patches with only minimal foreground markings, eliminating the need for extensive data training. The specific procedure of GrabCut \cite{GrabCut2004} is as follows:
\begin{itemize}
    \item Users provide an initial marking of the foreground object, which can be either a rectangular box or a set of pixels, indicating the approximate location of the foreground object. 
    \item GrabCut \cite{GrabCut2004} employs a Gaussian Mixture Model to represent the pixel distribution within the image and determines whether each pixel belongs to the foreground or background. The algorithm achieves segmentation results by minimizing the energy function $E$.
\end{itemize}
\begin{equation}
    E = \sum(-log_{2}P_{F_{i}}-log_{2}P_{B_{i}})
    \label{eq:grabcutE}
\end{equation}

As shown in Equation~\ref{eq:grabcutE}, $P_{F_{i}}$ and $P_{B_{i}}$ are the probabilities that pixel $i$ belongs to the foreground and background, respectively, as computed by the Gaussian Mixture Model. The algorithm outputs the segmentation results, identifying the image's foreground and background, once it reaches the maximum number of iterations or achieves convergence.

\subsection{Related Work} 

\subsubsection{Attacks}

Existing attack studies mainly target at computer vision tasks, e.g., image classification.The discovery of the adversarial attacks starts
from image classification tasks in the digital world \cite{mei23sur}. Szegedy et al. \cite{szegedy2013intriguing} propose the idea of generating adversarial examples with imperceptible perturbations to mislead the DNN-based image classifiers. Goodfellow et al. \cite{goodfellow2014explaining} propose FGSM to generate adversarial examples. Some improved FGSM methods are also proposed, such as IFGSM \cite{ifgsm2016adversarial} and PI-FGSM \cite{2020pifgsm}. Carlini and Wagner \cite{carlini2017towards} designed an optimization algorithm to generate adversarial examples with the least perturbations. 
As DNN-based systems are increasingly deployed in real-world scenarios, the rise of physical adversarial attacks has made strengthening their robustness both urgent and essential \cite{wang22sur}. Some studies focus on launching attacks in the physical world  \cite{sharif2016Access,sharif2019objectives,papernot2017practical,eykholt2018stopsign,brown2017advpatch}. 

The adversarial patch is mainly used to attack the object detection. Zhang et al. \cite{ZhangCPDSZ24} proposed to simulate the real environment in digital world attack, thus transfers adversarial textures seamlessly from digital to physical world. The adversarial medium includes patch and clothing \cite{wei24sur}. In the aforementioned attacks, \cite{thys2019fooling} is patch-medium attack, and \cite{xu2020T-shirt,wu2020cloak,komkov2019advhat,huang2020UPC} are the clothing-medium attacks.
Attacks on 3-D object detection systems have also been explored \cite{wang22sur, wang22fca}. In the future, we plan to research how to transfer our methods to detect 3-D attacks. We believe there is also the same semantics inconsistency in 3-D adversarial examples, i.e., the adversarial object can appear locally but disappear globally.

Except computer vision, there are some attacks on other tasks, e.g., online learning system \cite{vsrndic2014practical}, Google phishing pages classification \cite{liang2016cracking} and text classifier \cite{liang2018deep}.

\subsubsection{Defenses}

A lot of techniques have been proposed to detect adversarial examples. 

Some defenses focus on detecting the features of adversarial examples. Li et al. \cite{li2017statistics} utilize statistical features to distinguish adversarial examples from benign ones. Amirian et al. \cite{amirian2018trace} visualize the CNN features, track adversarial perturbations, and detect adversarial images based on statistical analysis of the feature responses. Liang et al. \cite{liang2021detecting} denoise the adversarial examples and compare the classification results for adversary detection. Xiao et al. \cite{Xiao_2018_ECCV} detect adversarial examples by checking the spatial inconsistency after applying two randomly selected patches to the given image. Lu et al. \cite{lu2017safetynet} propose SafetyNet to detect unreal scenes in an RGBD image by checking whether the image is consistent with its depth map. While different kinds of inconsistencies have been leveraged for adversary detection, in this paper, we focus on the semantics inconsistency that can be captured from the given input and robust enough to detect adversarial patched examples.
PatchZero \cite{xu2023patchzero} proposes a training scheme to generate a patch detector, which can defend the white-box patch attacks by repainting the detected adversarial pixels. This necessitates a substantial number of known adversarial examples, whereas our signature-based method does not; Additionally, our signature-independent method can defend the black-box attacks, without knowing any details about the adversarial examples.
Kim et al. \cite{kim2022APF} propose a defense method based on the feature of the abnormally high energy caused by adversarial patches, using layer-wise analysis to extract adversarial region proposals. 
Jing et al. \cite{jing2024pad} identify two inherent characteristics of adversarial patches, i.e. semantic independence and spatial heterogeneity, to localize and remove patches. 

Besides detecting adversarial images, efforts have been done to detect adversaries in other areas. Sun et al. \cite{sun2020towards} detect spoofed fake vehicles by checking the violations of physics, i.e., anomalously sensed free space and laser penetration. Choi et al. \cite{choi2018detecting} identify the physical attacks on robotic vehicles (RVs) by deriving and monitoring RVs' control invariants. Aghakhani et al. \cite{Aghakhani2018detecting} propose FakeGAN to detect deceptive customer reviews.

Improving the robustness of the DNNs has also been studied. Biggio et al. \cite{biggio2014security} provide a complete understanding of the classifier's behavior about adversarial examples, helping to improve the DNN's robustness with better design choices. Lyu et al. \cite{lyu2015unified} improve the robustness of the DNN to adversarial examples by considering the worst situation of perturbation and the gradient constraint in the model. Ba et al. \cite{18ba2014deep} and Hinton et al. \cite{19hinton2015distilling} extract knowledge from the DNN, with which they improve the resilience to adversarial examples. Papernot et al. \cite{papernot2016distillation} smooth the DNN model to reduce the enlargement of perturbations during feature extraction and make the model robust to defend small adversarial perturbations. Sharif et al. \cite{sharif2018suitability} suggest that other metrics like structural similarity \cite{wang2004image} should be considered together with Lp-forms to improve the model's robustness. Quiring et al. \cite{Prevent_Image-Scaling} reconstruct the image pixels to reveal adversarial pixels and filtering them out for classification correctness. Chen et al. \cite{chen2020pdf} build a robust PDF malware classifier with a new distance metric of operations on the PDF tree structure. Xiang et al. \cite{xiang2021patchguard} propose PatchGuard to defend the adversarial patch attacks on image classification. We have also demonstrated that our idea of region growing can effectively detect such attacks.
Ji et al. \cite{ji2021AD} propose a plugin trained on adversarial and benign examples that can be added to existing object detection models to detect adversarial patches. DIFFender \cite{kang2024diff} leverages text-guided diffusion models to detect and localize adversarial patches through Adversarial Anomaly Perception, integrating patch localization and restoration within a single framework. SAC \cite{liu2022sca} trains a robust patch segmenter through self-adversarial training and designs a shape completion algorithm to remove adversarial patches. Saha et al. \cite{saha2020role} demonstrate that limiting spatial context during object detector training improves robustness against adversarial patches. NAPGuard \cite{wu2024nap} uses the aggressive feature aligned learning and natural feature suppressed inference to enhance the detection capability.

Specifically, we perform comparison experiments on three defense methods, which are more relative to our methods. 
DetectorGuard \cite{xiang2021detectorguard} consists of a base detector and an objectness predictor, to detect the malicious mismatch of the patch-based hiding attack on object detection. Compared to it, our two defense methods exhibit higher accuracy. 
ObjectSeeker \cite{xiang2023objectseeker} detects the input image by dividing it into numbers of blocks and analyzing their detection results. Compared to ObjectSeeker, our signature-based defense approaches are with higher accuracy and faster detection, while the signature-independent defense approach exhibits higher accuracy and detection efficiency, with fewer detection steps per image on average.
Universal Defensive Frame (UDF)~\cite{yu2022udf} is optimized through competitive learning between detection threat and masking modules to defend person detection against adversarial patch attacks. The advantage of UDF is its universality, while in the cases that the adversarial object is small or the background is complex, its F1 score is lower than 0.5.

\section{Signature-based Detection}
\label{sec:3sig}

\subsection{Overview}
A natural way to detect the adversarial object is to find an effective rule (e.g., a signature) to distinguish the adversarial pixels in the input example and develop a mechanism to exclude them before feeding the example to the model. According to our observation, the adversarial patch actually possesses some distinct signatures to identify itself. 
We utilize the signatures to identify the possible adversarial patch in the input image, and then we filter out these identified pixels to generate a clean sample.

Classic saliency detection algorithms, e.g., Spectral Sesidual \cite{hou07sr} and Itti. et al \cite{itti98sm}, are first considered to localize this kind of signatures. However, in our preliminary experiment, they failed to separate any patches from objects in 168 adversarial samples generated from MS COCO~\cite{lin2014coco}. The results indicate that these two representative techniques are not effective in defending against attacks. We believe that not all saliency detection methods can be directly applied to the target problem. 

Across different scenarios, including both on the digital and physical world, we identify two exploitable signatures: Region Entropy and Guided Grad-CAM \cite{selvaraju2020grad}. We use these two signatures and design corresponding methods to filter out adversarial pixels in adversarial examples. The two signatures each possess their own distinct advantages. The former one is more suitable for digital world attacks, and the latter one is more suitable for physical world attacks.

\subsection{Signature: Region Entropy}

\begin{figure}[htbp]
    \centering
    \includegraphics[scale=0.35]{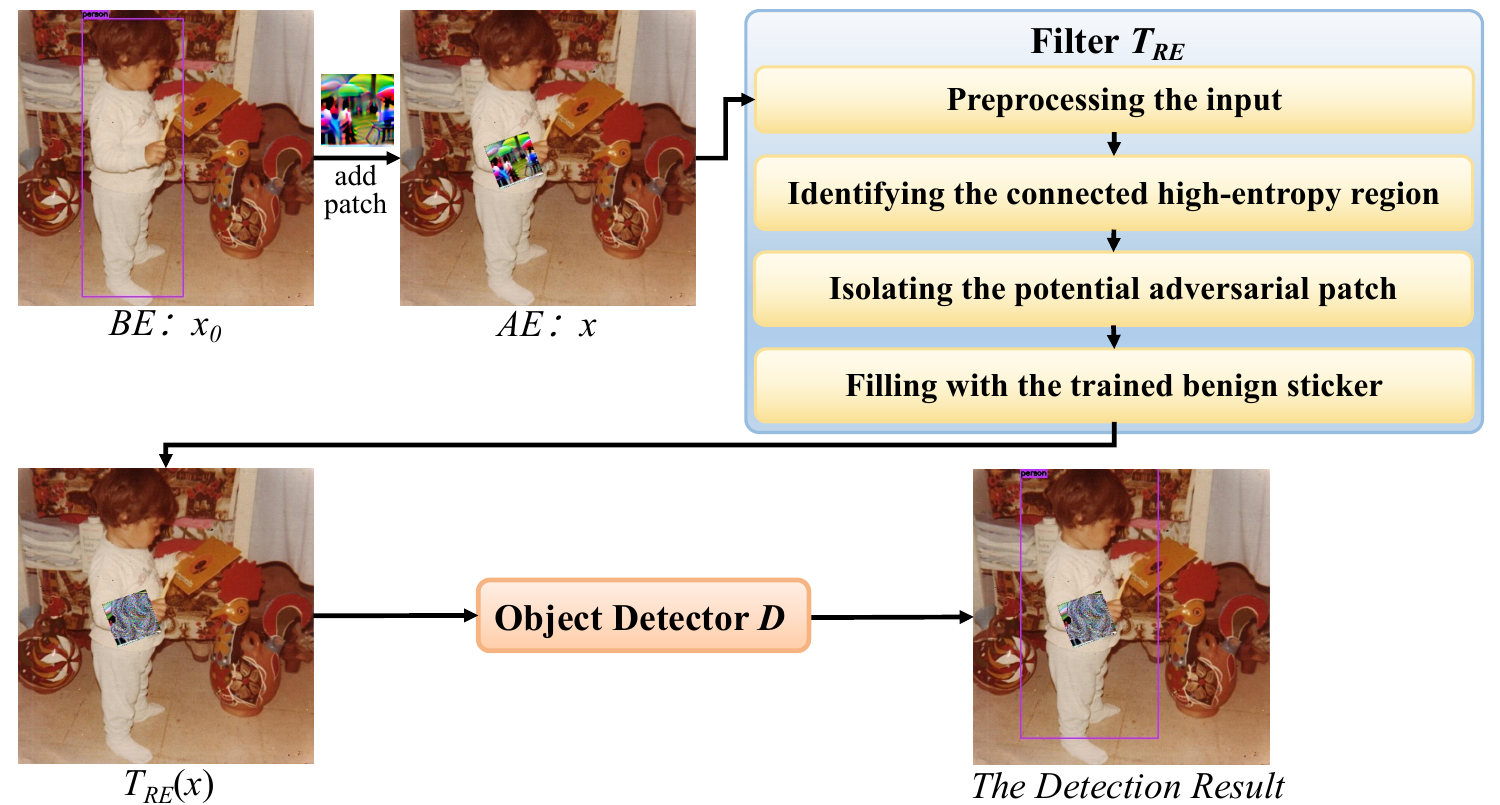}
    \vspace{-1mm}
    \caption{Workflow of the defense based on Region Entropy.}
    \label{fig:regionetropy-workflow}
    \vspace{-4mm}
\end{figure}

\begin{figure}
    \subfloat[input]{
    \includegraphics[width=0.15\linewidth]{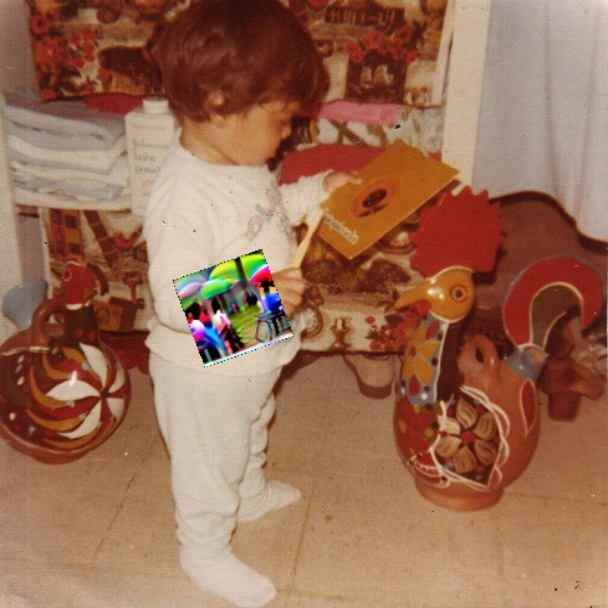}
    }
    \subfloat[]{
    \includegraphics[width=0.15\linewidth]{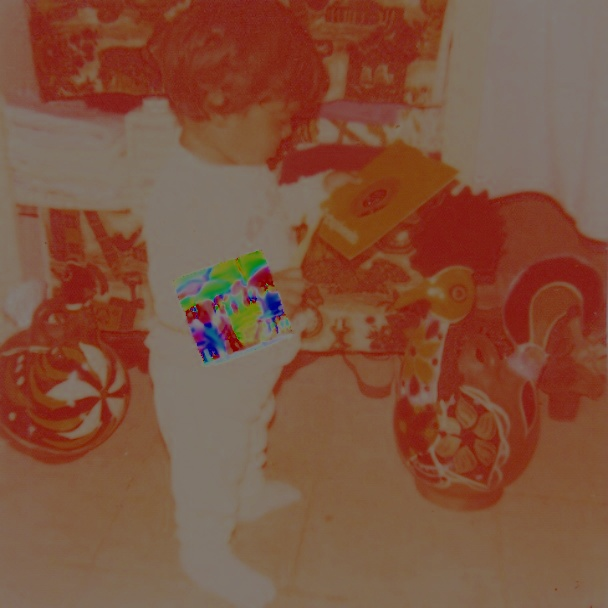}
    }
    \subfloat[]{
    \includegraphics[width=0.15\linewidth]{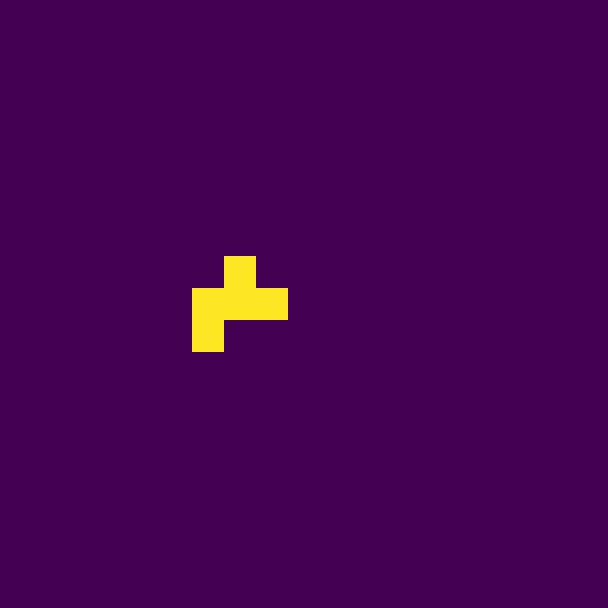}
    }
    \subfloat[]{
    \includegraphics[width=0.15\linewidth]{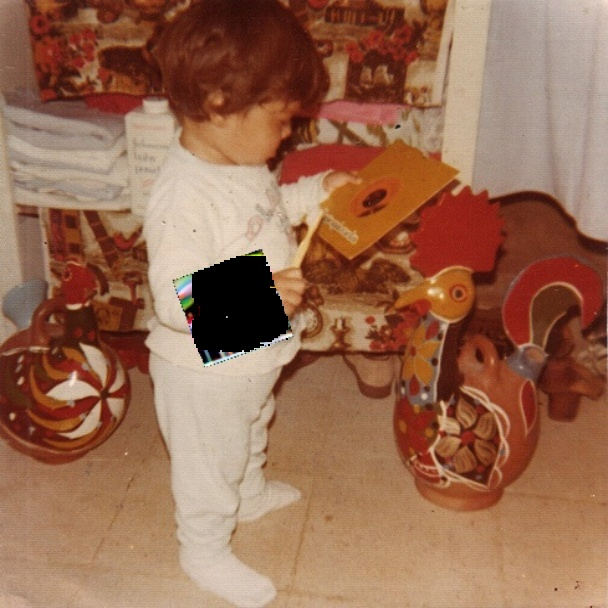}
    }
    \subfloat[]{
    \includegraphics[width=0.15\linewidth]{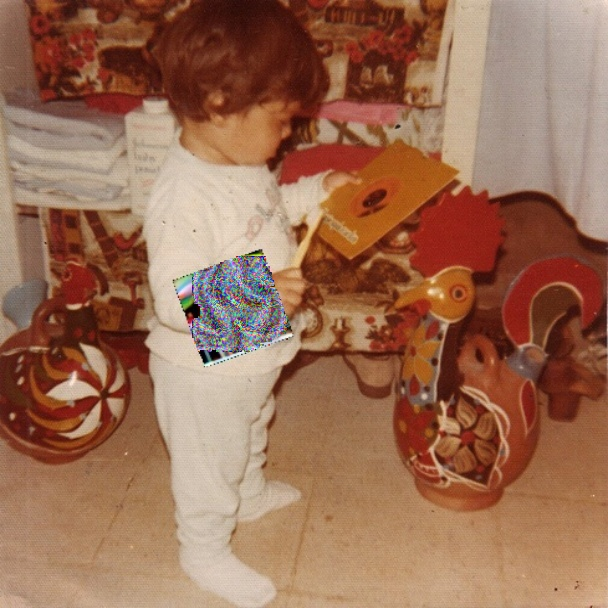}
    }
    \subfloat[detection]{
    \includegraphics[width=0.15\linewidth]{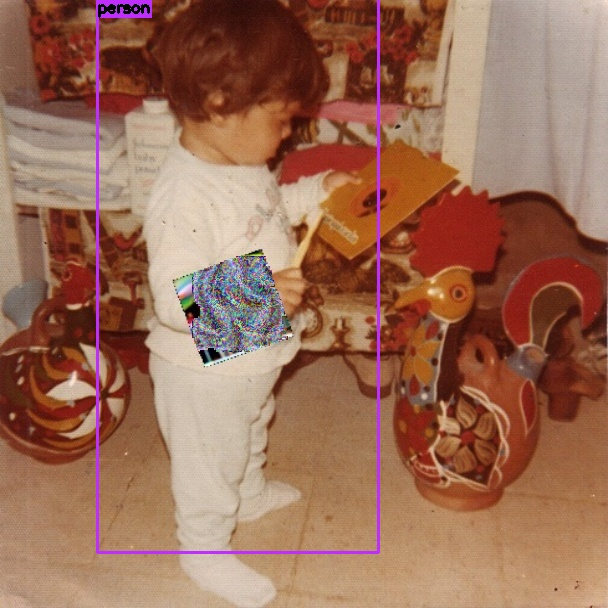}
    }
    \centering
    \caption{Processing steps of Region Entropy.}
    \label{fig:middleRE}
\end{figure}

In this study, adversarial patch regions are detected based on the intuition that the patch introduces high information quantity to the image, which can be measured by Region Entropy. We develop a method that uses Region Entropy for defense purposes. Figure~\ref{fig:regionetropy-workflow} shows the key components of the proposed Filter $T_{RE}$. The possible adversarial pixels of the input example $x$ are identified by Region Entropy, and we replace these identified pixels with a benign sticker to generate a filtered sample $T_{RE}(x)$. We initially aimed to directly identify adversarial patches by using Region Entropy. However, we found that relying solely on {Region Entropy} was not enough for accurately locating adversarial patches. To address this, we expand our approach by combining the use of {Region Entropy} with the image segmentation algorithm GrabCut \cite{GrabCut2004}. This allows us to achieve a more precise localization of the adversarial patches with low time cost. Next, we will introduce the steps of Filter $T_{RE}$ in detail. 
Figure \ref{fig:middleRE} displays the intermediate results, with (b) - (e) corresponding to the following four steps.

\textbf{Preprocessing the input image.}
To ensure accurate detection of adversarial patches, we normalize the brightness of input images to eliminate the detection effects caused by complicated brightness factor. The $T_{RE}$ converts input images to the HSV format, of which the V channel represents the brightness level, ranging from 0 to 255. The higher (lower) the V channel, the brighter (darker) the corresponding pixel. To reduce the impact of overly bright or dark areas in the images, we modify the V channel of each pixel to 128, no matter what the original value is. The other two channels (H and S) are left unchanged. This approach ensures that all images have a uniform brightness level, allowing our method to focus on the structural and color-based features of potential adversarial patches rather than being influenced by brightness variations.

\textbf{\textcolor{black}{Identifying} the connected high-entropy region within the input image.} 
For an input image, the Region Entropy is calculated for each block with Equation~\ref{eq:Block entropy}, in which $z$ is the number of pixels in the block and $p_i$ represents the probability of pixel value $i$ occurring in the block within a RGB color channel $c$. To calculate $p_i$, we first count the occurrences of each pixel value $i$ within the block, denoted as $count_i$. With the number of pixels $z =  (w_{img}/S) \times (h_{img}/S)$, where $w_{img}$ and $h_{img}$ correspond to the width and height of the input image (measured in pixels), we have $p_i = count_i/z$.
This ratio accurately reflects the frequency of occurrence for each specific pixel value within the block, thereby quantifying its probability distribution characteristics. 
By summing up the information entropy obtained for each color channel and applying Min-Max normalization to scale the result to [0, 1], we can obtain the final entropy result for the block.

\begin{equation}
    RE_{block \in S\times S}=\sum_{c \in \{R, G, B\}}\left(-\sum_{i=1}^{z}\left ( p_{i}\cdot log_{2}(p_{i}) \right ) \right)
    \label{eq:Block entropy}
\end{equation}

\begin{table}
    \centering
    \caption{Experiment results to determine a reasonable hyper-parameter $m$.}
    \label{tab:pyc_dfsm}
    \begin{tabular}{c|cccccccc}
    \hline
      $m = $   & $0.60$ & $0.65$ & $0.70$ & $0.75$ & $0.80$ & $0.85$ & \textbf{0.90} & $0.95$  \\ \hline
     Adv.  & 31 & 44 & 54 & 70 & 81 & 93 & 94 & 76 \\
      Benign  & 58 & 60 & 59 & 64 & 78 & 90 & 98 & 100 \\ \hline
     Total  & 89 & 104 & 113 & 134 & 159 & 183 & \textbf{192} & 176 \\ \hline
    \end{tabular}
\end{table}

For each block, if its normalized {Region Entropy} is greater than $m$, the block is considered a high entropy block. There is a trade-off for choosing a proper $m$. If $m$ is too small, the adversarial effect cannot be effectively suppressed, and the object can still be hidden by the patch. On the contrary, a large $m$ may result in benign blocks being processed and missing the object as well. To determine a reasonable hyper-parameter $m$, we conduct an experiment on the validation dataset.  
We randomly select 100 adversarial examples along with 100 benign examples of MS COCO \cite{lin2014coco} on YOLOv2 to form the validation dataset, and conducted a pilot study on the dataset using a range of hyperparameter values for $m$. The number of correctly identified examples across different parameter settings is summarized in Table~\ref{tab:pyc_dfsm}. It is evident that when $m=0.90$, the detection achieves the best overall performance, with accuracies of 94/100 for adversarial samples and 98/100 for benign examples. Therefore, unless otherwise stated, we use $m=0.90$ in all subsequent experiments.

\begin{figure}[htbp]
    \centering
    \includegraphics[scale=0.25]{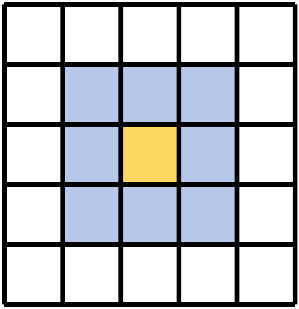}
    \caption{Eight-connectivity diagram.}
    \label{fig:eight-connected}
\end{figure}

We first filter high-entropy blocks based on $m$. Subsequently, the connected regions are computed using a depth-first search (DFS) algorithm, considering 8-connectivity rules in a 3$\times$3 grid. The connectivity analysis itself is independent of the block scores, as the high-entropy blocks have already been identified in the prior step. As illustrated in Figure~\ref{fig:eight-connected}, the blue blocks surrounding the yellow block represent the connected region of the yellow block in the context of 8-connectivity. 

\textbf{Isolating the potential adversarial patch.}
We have already introduced the principles of the GrabCut algorithm \cite{GrabCut2004} in Section 2. In this study, we aim to segment the adversarial patch by applying the GrabCut algorithm. To initiate this process, the minimum bounding rectangle of the largest connected region identified in the previous step is first used to define the foreground object. This rectangle serves as the input to GrabCut, which then outputs the isolated potential patch.
In addition, to accelerate the speed of image segmentation, we employ a downsampling strategy. Taking YOLOv8 as an example, we downscale the image from $640\times 640$ pixels to $320\times 320$ pixels before performing the segmentation. This downsampling approach maintains segmentation effectiveness while significantly improving detection speed.

\begin{figure}[htbp]
    \centering
    \includegraphics[scale=0.4]{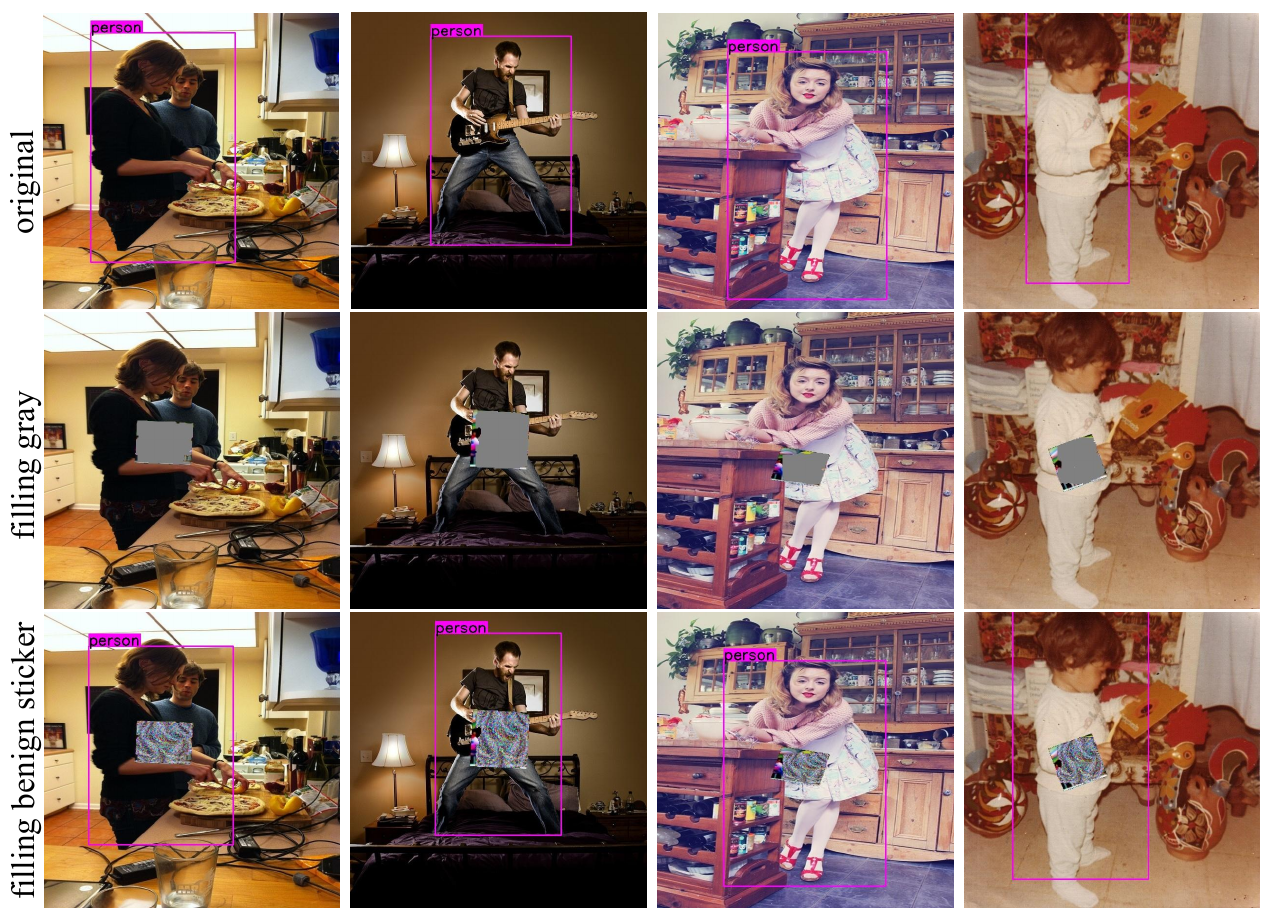}
    \vspace{-1mm}
    \caption{Different fillings result in different detection results.}
    \label{fig:sticker vs gray}
    \vspace{-4mm}
\end{figure}

\textbf{Filling with the trained benign sticker.} 
To minimize the impact of adversarial patches on the image while preserving object detection performance, we introduce a specially trained benign sticker. Unlike simplistic solid color fillings that degrade detection accuracy even the adversarial pixels are removed (see Figure~\ref{fig:sticker vs gray}), the benign sticker is trained to ensure the detection results align closely with that of the benign unaltered images, including $xywh$, $object$ $probability$ and $class$ $score$. The benign sticker is trained only on benign examples. The training process didn't involve any adversarial information. We trained the sticker on the MS COCO dataset with the YOLOv2 model. The loss is formulated as Equation~\ref{eq:loss of sticker}. 
\begin{equation}
\begin{split}
     Loss &=MSE(xywh,xywh')\\
     &+MSE(obj\_prob,obj\_prob') \\
     &+MSE(cls\_score,cls\_score')
\end{split}
\label{eq:loss of sticker}
\end{equation}
Here, $xywh$, $obj\_prob$ and $cls\_score$ represent the bounding box parameters, object confidence, and class scores of the original (benign) image, respectively, while $xywh'$, $obj\_prob'$ and $cls\_score'$ denote the corresponding metrics for the image with the sticker applied on the benign object. To ensure robustness, augmentations like rotation and scaling are applied, making the sticker adaptable to varying orientations and sizes of the adversarial patch regions.  

We performed an experiment on the adversarial examples from the validation dataset, which was used to determine $m$ above, to prove that our benign sticker is much better for filling than gray pixels. The detection accuracy of the gray filling is 89\%, while the accuracy of our benign sticker filling is 96\%. Figure~\ref{fig:sticker vs gray} presents some results that different fillings have different effects on detection results. In the second row, the patches are filled with gray pixels, resulting in the attacked object remaining undetected. In the bottom row, the patches are filled with our benign sticker, allowing the attacked object to be detected.

\subsection{Signature: Guided Grad-CAM}
\begin{figure}[htbp]
    \centering
    \includegraphics[scale=0.4]{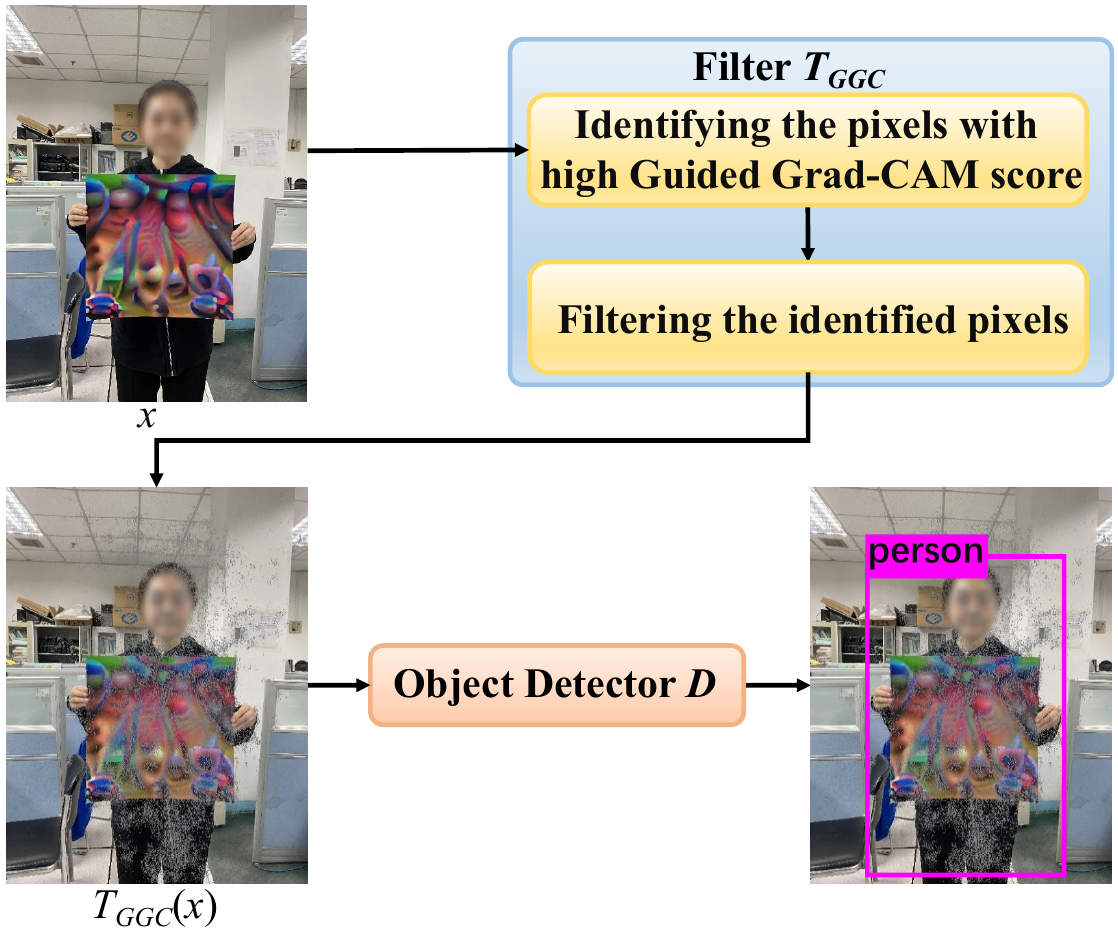}
    \vspace{-1mm}
    \caption{Workflow of the defense based on Guided Grad-CAM.}
    \label{fig:filter}
    \vspace{-4mm}
\end{figure}

In addition to the signature {Region Entropy} mentioned above, we also employ another signature known as Guided Grad-CAM. We propose a defense based on Guided Grad-CAM called the Filter $T_{GGC}$. As shown in Figure~\ref{fig:filter}, the key component of our method is a filter $T_{GGC}$. An example $x$ will be denoised by $T_{GGC}$ to generate a filtered sample $T_{GGC}$($x$) as follows. Next, we will introduce the two steps of $T_{GGC}$ in detail.

\textbf{Identifying the adversarial pixels by Guided Grad-CAM.} Guided Grad-CAM \cite{selvaraju2020grad}, a highly class-discriminative visualization method, is employed to highlight the potential important input pixels for certain model objectives. Each input pixel is assigned a quantitative importance estimation and we term it the \textit{Guided Grad-CAM score}. In a preliminary experiment involving adversarial samples targeting YOLOv2, we observed that the adversarial patch pixels typically yielded significantly higher Guided Grad-CAM scores for the Person class compared to other input pixels. In other words, the pixels of the adversarial patch are likely to contribute more to the Person output, rather than fewer, as we initially expected. In this study, we utilize the Guided Grad-CAM scores calculated for the Person class for filtering. The Guided Grad-CAM score for the output \textit{Person} class probability is calculated as follows. First, the gradient of the \textit{Person} class probability concerning the output layer feature maps is computed by Equation~\ref{equ:grad}:
\begin{equation}
    g_{k}^{p} = \frac{\partial P_{Person}}{\partial F^{k}}, 
    \label{equ:grad}
\end{equation}
where $P_{Person}$ is the probability of the \textit{Person} class, and $F^{k}$ is the $k$-th output layer feature map. Second, as shown in Equation~\ref{equ:average}, we use the global average pooling to get the neuron importance weights,
\begin{equation}
    \alpha _{k}^{p} = \frac{1}{H \cdot W}\sum_{i}^{H}\sum_{j}^{W} g_{k}^{p}(i, j),
    \label{equ:average}
\end{equation}
where ($H$, $W$) are the height and width of $F^{k}$. Third, the coarse Grad-CAM score $L^{p}$ is computed with Equation~\ref{equ:L},
\begin{equation}
    L^{p} = ReLU \left( \sum_{k} \alpha _{k}^{p}F^{k} \right).
    \label{equ:L}
\end{equation}
$L^{p}$ is upsampled to the input image resolution using bilinear interpolation, and finally it is multiplied by $\nabla_{x}P_{person}$ to get the Guided Grad-CAM scores for each input pixel, where $x$ is the input image. We choose the \textit{Person} class as the target class in this study. The target class probability $P_{Person}$ can be collected in the last layer, and the Guided Grad-CAM scores can be easily computed as done in \cite{selvaraju2020grad}. 

\begin{table}[htbp]
    \centering
    \setlength{\tabcolsep}{4pt}
    \caption{Experiment to decide hyper-parameter \textit{n}.}
    \label{tab:Validation n}
    \begin{tabular}{c|cccccccccc}
        \hline
        $n = $ & $0.5\%$ & $1\%$ & \textbf{2\%} & $3\%$ & $4\%$ & $5\%$ & $6\%$ & $7\%$ & $8\%$ & $9\%$ \\ \hline
        Adv. & 41 & 52 & 57 & 55 & 50 & 48 & 43 & 43 & 45 & 45 \\ 
        Benign  & 97 & 94 & 92 & 87 & 85 & 85 & 82 & 74 & 74 & 67 \\ \hline 
        Total  & 138 & 146 & \textbf{149} & 142 & 135 & 133 & 125 & 117 & 119 & 112 \\ \hline 
    \end{tabular}
\end{table}

\textbf{Filtering the identified pixels.} For each input image, its top $n$ pixels with high Guided Grad-CAM scores are identified and filtered. The identified pixels are colored with a fixed value to erase their adversarial effect and generate $T_{GGC}$($x$). Taking $T_{GGC}$($x$) as the input, the object detector $D$ can then effectively detect the object hidden by the adversarial patch. There is a trade-off for choosing a proper $n$. If $n$ is too small, the adversarial effect cannot be effectively suppressed, and the patch can still hide the object person. On the contrary, a large $n$ may result in the benign pixels being filtered and missing the object person as well.

A practical experiment is conducted on the validation dataset to identify an appropriate $n$. As shown in Table \ref{tab:Validation n}, the optimal performance of the proposed method is attained when n = 2\%, which we use in subsequent experiments.

Several different coloring schemes have been tested. According to our experiments, coloring them in gray works better than other colors, e.g., black, white, or random pixels, as shown in Table \ref{tab:different filtering colors}. Note that, some non-patch pixels may also possess high Guided Grad-CAM scores. Filtering them may remove some details of the target person. Fortunately, in most cases, the object detection model can successfully detect the target person even from a small part of her or him. Filtering some benign pixels generally does not significantly affect the model's detection performance.

\begin{table}[htbp]
    \centering
    \caption{Validation experiment to decide filtering color.}
    \label{tab:different filtering colors}
    \begin{tabular}{c|cccc}
    \hline
       Coloring Schemes  & black & white & random & \textbf{gray} \\ \hline
        Adversarial Examples & 11 & 53 & 30 & 57 \\
        Benign Examples & 30 & 70 & 61 & 92 \\ \hline
        Total Examples & 41 & 123 & 91 & \textbf{149} \\ \hline
    \end{tabular}
\end{table}

\subsection{Evaluation}
\label{subsec:3eva}

\begin{table*}[htbp]
\centering
\caption{The detection results of three comparative methods and our methods.}
\label{table:3 detection evaluation}
\resizebox{\textwidth}{!}{
\begin{tabular}{cccccccccccccccccc}
\hline
& \multirow{2}{*}{Model} & \multirow{2}{*}{Dataset} & \multicolumn{3}{c}{DetectorGuard} & \multicolumn{3}{c}{ObjectSeeker} & \multicolumn{3}{c}{\textcolor{black}{UDF}} & \multicolumn{3}{c}{Region Entropy} & \multicolumn{3}{c}{Guided GradCAM} \\ \cline{4-18}
&     &     &  P & R & F1  &  P & R & F1 &  P & R & F1 &  P & R & F1 &  P & R & F1 \\ \hline
\multirow{8}{*}{Digital} & \multirow{2}{*}{YOLOv2} & COCO & 0.487 & 0.435 & 0.459 & 0.353 & 0.286 & 0.316 & 0.944 & 0.405 & 0.567 & 0.988 & 0.952 & \textbf{0.970} & 0.855 & 0.595 & \textbf{0.702} \\ 
                   & & VOC & 0.441 & 0.333 & 0.380 & 0.351 & 0.289 & 0.317 & 0.579 & 0.244 & 0.344 & 0.933 & 0.933 & \textbf{0.933} & 0.757 & 0.622 & \textbf{0.683}  \\ 
& \multirow{2}{*}{YOLOv4} & COCO & 0.636 & 0.233 & 0.341 & 0.909 & 0.333 & 0.488 & 0.957 & 0.733 & 0.830 & 1.000 & 0.867 & \textbf{0.929} & 1.000 & 0.733 & \textbf{0.846}\\ 
                   & & VOC & 0.500 & 0.241 & 0.326 & 0.571 & 0.123 & 0.203 & 0.941 & 0.552& 0.696 & 1.000 & 0.931 & \textbf{0.964} & 0.913 & 0.724 & \textbf{0.808} \\ 
& \multirow{2}{*}{YOLOR} & COCO & 0.143 & 0.093 & 0.112 & 1.000 & 0.870 & 0.931 & 1.000 & 0.623 & 0.767 & 1.000 & 0.943 & \textbf{0.971} & 0.760 & 0.358 & 0.487 \\ 
                   &  & VOC & 0.429 & 0.286 & 0.343 & 0.963 & 0.619 & 0.754 & 1.000 & 0.359 & 0.528 & 0.973 & 0.923 & \textbf{0.947} & 0.880 & 0.564 & 0.688  \\ 
& \multirow{2}{*}{YOLOv8} & COCO & 0.823 & 0.294 & 0.433 & 0.684 & 0.353 & 0.466 & 1.000 & 0.683 & 0.812 & 0.995 & 0.910 & \textbf{0.950} & 0.989 & 0.407 & 0.577  \\ 
                   &  & VOC & 0.662 & 0.288 & 0.401 & 0.661 & 0.372 & 0.476 & 0.980 & 0.625 & 0.763 & 0.983 & 0.931 & \textbf{0.957} & 0.961 & 0.384 & 0.549 \\ \hline
\multirow{4}{*}{Physical} & \multirow{4}{*}{YOLOv2}  & Indoor & 0.488 & 0.920 & 0.637 & 0.496 & 0.410 & 0.449 & 1.000 & 0.090 & 0.165 & 0.945 & 0.747 & \textbf{0.834} & 1.000 & 0.880 & \textbf{0.936}  \\ 
                   &  & Outdoor & 0.448 & 0.677 & 0.539 & 0.671 & 0.313 & 0.427 & 1.000 & 0.467 & 0.636 & 0.914 & 0.526 & \textbf{0.668} & 0.975 & 0.893 & \textbf{0.932}  \\  
                   &   & \textcolor{black}{MyAdvTshirt}  & 1.000 & 0.111 & 0.200 & 0.864 & 0.264 & 0.404 & 0.982 & 0.750 & 0.850 & 1.000 & 0.917 & \textbf{0.957} & 0.983 & 0.806 & \textbf{0.886}  \\
                  &   & AdvTshirt \cite{xu2020T-shirt} & 0.500 & 0.042 & 0.077 & 0.818 & 0.750 & 0.783 & 0.706 & 0.833 & 0.764 & 0.941 & 0.727 & \textbf{0.821} & 1.000 & 1.000& \textbf{1.000}  \\ \hline
\multicolumn{3}{c}{Total} & 0.564 & 0.488 & 0.463 & 0.620 & 0.372 & 0.456 & 0.973 & 0.463 & 0.593 & 0.964 & 0.814 & \textbf{0.876} & 0.949 & 0.643 & \textbf{0.745}   \\ \hline 
\end{tabular}}
\end{table*}

\begin{figure*}[htbp]
    \centering
    \includegraphics[width=0.9\textwidth]{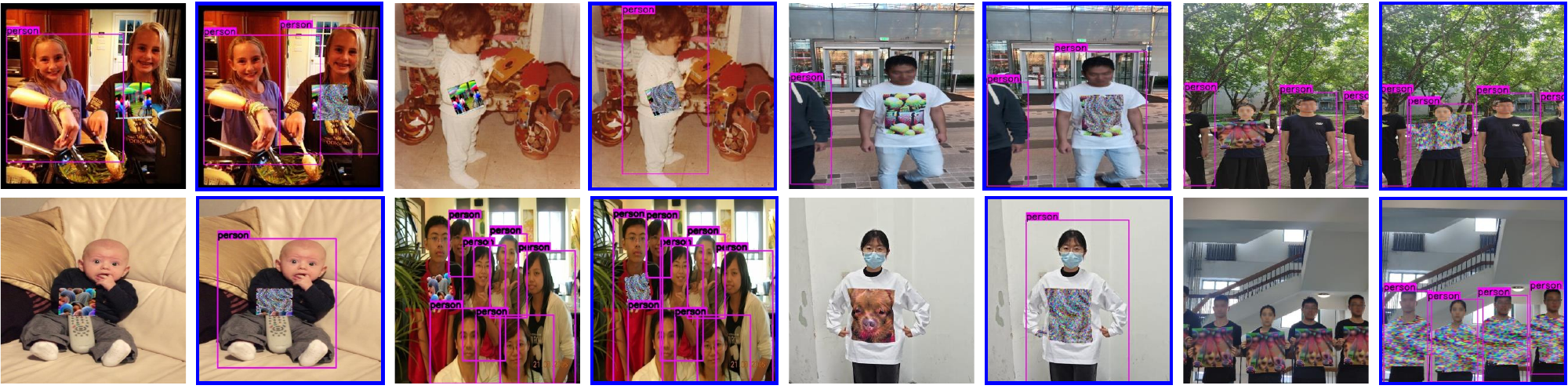}
    \caption{\revbegin{2}{1} \nnewljc{The defense examples on digital and physical datasets, based on Region Entropy.} \revend{2}{1}
    }
    \label{fig:public_result_region_entropy}
\end{figure*}

\begin{figure*}[htbp]
    \centering
    \includegraphics[width=0.9\textwidth]{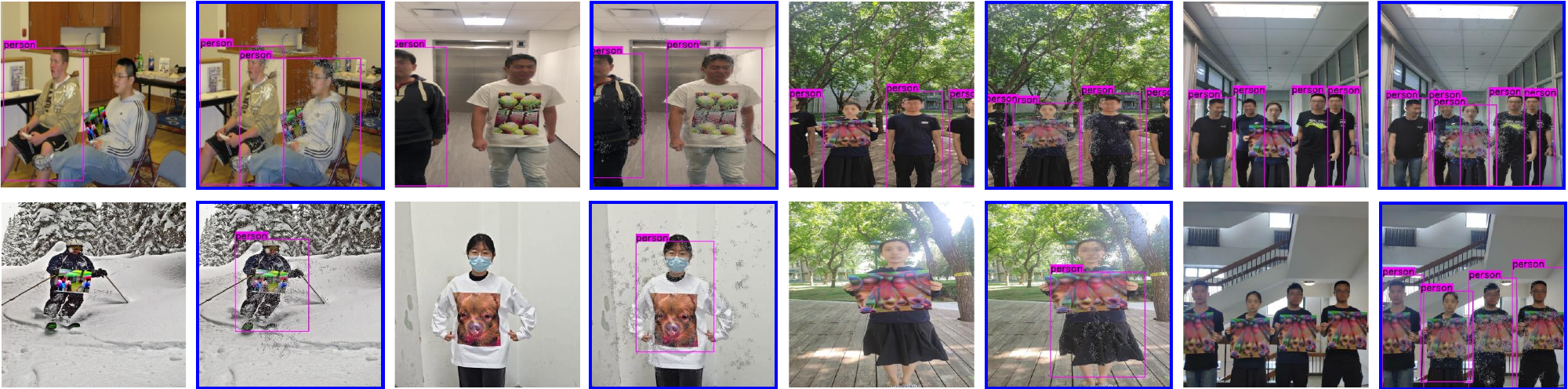}
    \caption{\revbegin{2}{1} \nnewljc{The defense examples on digital and physical datasets, based on Guided Grad-CAM.} \revend{2}{1}
    }
    \label{fig:public_result_ggc}
\end{figure*}

\subsubsection{Experiment settings} 
\textbf{Datesets.} AdvPatch \cite{thys2019fooling} attack is a classical work that targets hiding person objects.  Given that the authors provided their well-trained adversarial patch, we utilize it to generate adversarial examples on two public datasets, i.e., MS COCO \cite{lin2014coco} and PASCAL VOC \cite{everingham2010voc}. 
We also generate adversarial patches for Scaled-YOLOv4 \cite{scaled-yolov4}, YOLOR \cite{YOLOR} and YOLOv8 \cite{yolov8} based on the source code provided by AdvPatch, and then generate adversarial examples on COCO and VOC.
We use pairs of adversarial and benign examples, with a total of 905 adversarial examples and 905 corresponding benign examples, for a combined total of 1810 examples on COCO and VOC. Only adversarial examples that successfully attack the object detection models are included, while failed examples are discarded.
The datasets also contain 24 physical adversarial examples that can successfully attack YOLOv2, from the AdvTshirt attack \cite{xu2020T-shirt}.

We also generated two other kinds of adversarial examples in physical world.
First, we take hundreds of examples of the attacker holding a printed patch by AdvPatch, from which we select those that could successfully evade object detection. Correspondingly, we take photos without adversarial patches in the same scenes, serving as benign examples. These examples include single and crowd scenes, indoor and outdoor. The number of such physical world examples is 1200. 
Second, to demonstrate that our method is effective not only against easily locatable patches but also against adversarial examples where patches are integrated into clothing, we printed the adversarial pattern \cite{huang2020UPC} on T-shirts, called \textit{MyAdvTshirt}. We tried our best to obtain 72 such adversarial examples targeting YOLOv2 and 72 corresponding benign examples.

\textbf{Models and Metrics.}
The detection models used in our work employ their default or commonly used thresholds, specifically, 0.5 for YOLOv2, 0.4 for YOLOv4 and YOLOR, and 0.25 for YOLOV8.
In this paper, when applying our proposed method, we consider an object to be successfully detected when the confidence level exceeds the model's threshold. Conversely, if the detection result falls below the model's threshold, we consider that the object was not successfully detected. 
We performed our experiments on one GPU of GeForce RTX 2080 Ti.

The metrics measure the actual performance of person detection after applying our defense methods. Detection is considered correct in two conditions: (1) \textit{True Positive}: the model identifies all the hidden objects in an adversarial example, or (2) \textit{True Negative}: the model does not miss any objects that can be correctly identified in a benign example.
Violating either condition will be treated as an unsuccessful defense, i.e., \textit{False Negative} when violating (1) and \textit{False Positive} for (2).
The precision (P), recall(R) and the F1 score (F1) are then computed to evaluate the effectiveness of our methods in defending against adversarial attacks while preserving the utility of person objects without adversarial patches.

\begin{table}[htbp]
    \centering
    \caption{The detailed detection results on our physical dataset, based on Guided Grad-CAM.}
    \label{tab:ggc mydataset detailed}
    \begin{tabular}{ccccc}
    \hline
         \multicolumn{2}{c}{Data Set} & Precision & Recall & \textbf{F1 score}   \\ \hline
         \multirow{2}{*}{Indoor AdvPatch}         
         & single &  1.000  & 0.930 & \textbf{0.964}   \\ 
         & crowd  &  1.000  & 0.855 & \textbf{0.922}   \\ \hline
         \multirow{2}{*}{Outdoor AdvPatch}
         & single  &  1.000  & 0.920 & \textbf{0.958}   \\  
         & crowd  &  0.962  & 0.880 & \textbf{0.919} \\ \hline
         \textcolor{black}{My AdvTshirt} & single & 0.983 & 0.806 & \textbf{0.886} \\ \hline
         Indoor AdvTshirt & crowd    &  1.000 & 1.000  & \textbf{1.000}   \\ \hline
         Outdoor AdvTshirt & crowd  &  1.000 & 1.000  & \textbf{1.000}  \\ \hline       
    \end{tabular}
\end{table}

\subsubsection{Detection results} 
The results of both signatures are shown in Table~\ref{table:3 detection evaluation}.

Regarding the detection results of {Region Entropy} shown in Table \ref{table:3 detection evaluation}, the precision for all datasets on each model exceeds 0.9, with many nearly reaching 1.0. 
The recall is above 0.910, except for the COCO dataset using YOLOv4, which is 0.867.
The recall for most items ranges from 0.923 to 0.952. The F1 score for all items ranges from 0.929 to 0.971. We also perform a defense experiment based on {Region Entropy} on the physical world dataset, with $m=0.8$ and the four-connectivity method. Results indicate that the {Region Entropy} signature also possesses a certain defense capability against adversarial samples in the physical world, with precision above 0.9. Through further analysis, we find that the scenario with a high false negative rate is a crowd scene, especially an outdoor crowd scene, which is too complicated to detect accurately. On AdvTshirt dataset, our region-entropy-based method also has certain detection capability. Figure \ref{fig:public_result_region_entropy} displays adversarial examples and their detection results. 
The method also achieves a high F1 performance of 0.957 on MyAdvTshirt, demonstrating the method's resilience against wearable adversarial patterns.

Guided Grad-CAM demonstrates a bit less effectiveness compared to Region Entropy; however, Guided Grad-CAM delivers overall better performance on the physical dataset. Table~\ref{tab:ggc mydataset detailed} shows the breakdown results for each physical scenario. The F1 scores in 6 out of 7 cases are above 0.9. Except that the outdoor crowd scenes and MyAdvTshirt obtain near perfect \nnewljc{precision}, all others reach the perfect \nnewljc{precision}. It is also worth noting that, the defense capability of the Guided Grad-CAM signature against the AdvTshirt dataset reaches 100\%. Figure \ref{fig:public_result_ggc} presents some examples of defending against the attacks with the Guided Grad-CAM-based method.

\begin{table*}[htbp]
    \centering
    \caption{Average time cost (\textit{ms}) of the signature-based methods.} 
    \label{tab:time cost public}
    \resizebox{\textwidth}{!}{
    \begin{tabular}{ccccccccccccc}
    \hline
        & \multicolumn{6}{c}{YOLOv2} & \multicolumn{2}{c}{YOLOv4} & \multicolumn{2}{c}{YOLOR} & \multicolumn{2}{c}{YOLOv8} \\ \cline{2-13}
         & COCO & VOC & Indoor & Outdoor & \textcolor{black}{MyAdvTshirt} & AdvTshirt & COCO & VOC & COCO & VOC & COCO & VOC\\ \hline
    Region Entropy & 151.7 & 151.7 & 168.0 & 170.0 & 142.5 & 176.6 & 81.8 & 81.8 & 594.0 & 534.6 & 171.6 & 159.4\\ 
    Guided Grad-CAM & 144.6 & 147.9 & 92.4 & 92.4 & 137.5 & 167.0 & 120.9 & 123.8 & 281.6 & 273.6 & 85.5 & 81.5 \\ \hline
    \end{tabular}}
\end{table*}

Despite the defense performance, the two kinds of signatures have their own advantages in different scenarios. Specifically, Region Entropy is more appropriate for the digital world, where the public dataset is used. Even in complex scenarios, including crowds and cluttered backgrounds, Region Entropy can accurately locate the adversarial patches in adversarial examples. On the contrary, Guided Grad-CAM could not accurately identify adversarial patches in examples with complicated backgrounds, resulting in unsatisfied performance. However, Guided Grad-CAM is better suited for the physical world. In physical world, the light and color of the adversarial patch are significantly different from those in the digital world. Guided Grad-CAM can still recognize the adversarial patch, while the recognition capability of the Region Entropy signature is reduced. 

The differences arise because Region Entropy, based on the image's color, extracts shallow features and is effective when the adversarial patch has minimal color distortion. In contrast, Guided Grad-CAM relies on the neural network to extract deep features, allowing it to be effective even with high color distortion.

\subsubsection{Real-time detection ability}
\label{subsubsec:Real-time detection ability}
Table~\ref{tab:time cost public} shows the time cost of the Region-Entropy-based detection and Guided-Grad-CAM-based detection. 
We calculated the average detection time for each detection model. The results demonstrate that YOLOv2, YOLOv4, and YOLOv8 can achieve real-time performance, with only 160ms, 80ms, and 165ms per detection, respectively. YOLOR demands a higher computation time of 560ms on average, which can be attributed to its larger sample size, nearly four times that of other models. In most scenarios, 6-7 frames per second or more can be detected, satisfying the requirement of real-time detection.

\begin{newljc}{black}
\revbegin{1}{3} For the signature-based method, the detection process is consistent for both single scenes and crowd scenes, as each sample is processed only once. However, our investigation shows that, the average detection time for crowd scenes is 16.7\% higher than that for single scenes. This difference is due to the higher information density in crowd scenes, which leads to an increased computation time during feature processing. \revend{1}{3}
\end{newljc}

\subsubsection{Comparison with other methods}
\label{subsubsec: 3comparison}

We conducted comparison experiments using the well-known methods DetectorGuard \cite{xiang2021detectorguard}, ObjectSeeker \cite{xiang2023objectseeker}, and UDF \cite{yu2022udf}, applied them to all the detection models and datasets we used. 
Our experiments followed their paper and the associated source code provided by the authors.
Table~\ref{table:3 detection evaluation} shows the comparison results. The overall performance (F1) of Region Entropy outperforms all the three methods on all models and datasets, while Guided Grad-CAM outperforms DetectGuard, ObjectSeeker and UDF in 12,10 and 9 cases, respectively, out of all the 12 cases.

\subsubsection{Detection on defense-aware attack}
\label{sec:3sig-adaptive}
We assume that adaptive attackers aim to evade our proposed signature-based detection. That is, they may generate adversarial patches that not only bypass object detection but also diminish their own {Region Entropy}. We obtained 78 adaptive adversarial examples on COCO that successfully evaded object detection YOLOv2. Our method manages to detect 76.9\% of these adaptive examples.

To enhance our defense against adaptive attacks, we incorporated the Guided-Grad-CAM-based detection, to defend together with the region-entropy-based method. An attacker aiming to evade our detection must bypass both the high {Region Entropy} signature and the high Guided Grad-CAM signature.
Armed with this strategy, we are able to successfully detect 91.0\% of the adaptive adversarial examples. The precision of benign examples is slightly reduced because we only consider a benign example correctly detected when both detection methods agree. Table~\ref{tab:defense aware result} shows the detection results in detail. Mode 1 represents only Region Entropy; Mode 2 represents only Guided Grad-CAM ; Mode 3 represents both Region Entropy and Guided Grad-CAM.

\begin{table}[htbp]
    \centering
    \caption{The detection results of defense-aware attack.}
    \label{tab:defense aware result}
    \begin{tabular}{c | cccc}
    \hline
      Mode   & \nnewljc{Precision} & Recall & \textbf{F1 score} & Avg. Time($ms$) \\ \hline
     {Mode 1} & 1.000  & 0.769 & \textbf{0.870} & 151.7 \\
     {Mode 2} & 0.800 & 0.410 & \textbf{0.542} & 144.6\\
     {Mode 3} & 0.899 & 0.910 & \textbf{0.904} & 296.3 \\ \hline
    \end{tabular}
\end{table}

We also generate extra adaptive adversarial patches for both signatures. Nevertheless, because of the gradient conflict among various losses, obtaining such an extreme adversarial patch is difficult.
Using the COCO dataset, we only procure 23 adversarial examples that could successfully bypass object detection while simultaneously reducing both signatures. Of these 23 samples, 18 are still unable to circumvent the detection of both signatures, indicating that our new strategy is highly robust against adaptive attacks.

\subsubsection{Causes of FPs and FNs}
\begin{figure}[htbp]
  \centering
  \subfloat[An FP case study \label{fig:sig_based_FP}]{
        \includegraphics[width=0.23\textwidth]{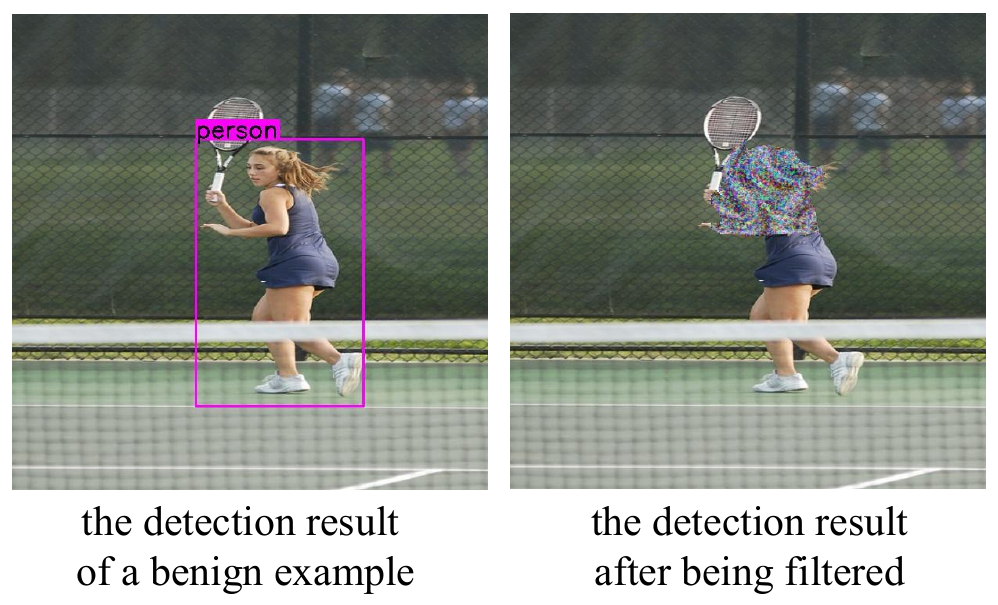}
        } %
  \subfloat[An FN case study \label{fig:sig_based_FN}]{
        \includegraphics[width=0.23\textwidth]{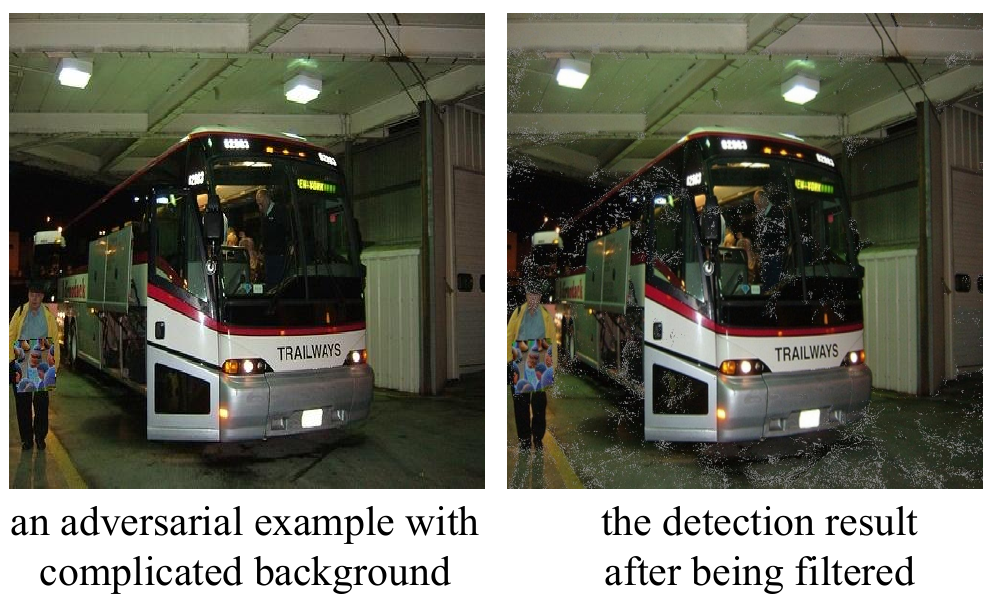}
        } %
  \caption{An FP and an FN of the signature-based method.}
\end{figure}

For the signature-based method, the main cause of false positives (FPs) is that the signatures of benign target persons are mistakenly detected as adversarial patch features. During the detection of benign examples, the key features of the original target person may be inadvertently filtered out, leading to detection failures, as shown in Figure \ref{fig:sig_based_FP}.

The primary cause of false negatives (FNs) is related to the specific features used in our method. In certain scenarios, such as those with overly complex backgrounds, the distinct features of the patch itself may become less prominent. As a result, when filtering features in adversarial samples, the method may fail to successfully remove adversarial pixels, leading to an inability to defend against the attacks. An example is shown in Figure \ref{fig:sig_based_FN}.

\subsubsection{Detection on other object classes}
\label{subsub:other obj cls}
We specifically generated adversarial examples on \textit{Planes} and \textit{Cats} using the COCO dataset. We apply YOLOv8 on COCO-based adversarial examples and the corresponding benign examples. For Region Entropy, the F1 scores of \textit{Planes} and \textit{Cats} are 0.957 and 0.912, respectively. For Guided Grad-CAM, the F1 scores are 0.765 and 0.640, respectively. The results demonstrate the applicability of our methods to object classes other than the person.

\section {Signature-Independent Detection}
\label{sec:defense_growth}

\subsection{Overview}
\label{subsec:growth_overview}

As presented in Section~\ref{sec:3sig}, the signature-based defense method is proven to be effective against the existing attacks. However, the employed signature is attack-specific, and we demonstrate the potential risk in Section~\ref{sec:3sig-adaptive}.

To defend the upgraded adversarial patch generation algorithm and other unknown sophisticated attack techniques, we propose a general and signature-independent detection method in this section. The proposed method is based on an internal semantics structure rather than some kind of mutable superficial characteristics. In fact, the local and global content  semantics are consistent in benign examples but inconsistent in adversarial ones. If a benign object can be detected from a part of the input image, it will certainly be detected from a larger one. In other words, if an object can be detected locally, it should be detected globally. However, an adversarial object can become disappeared in the larger part containing enough adversarial patch pixels. This semantics structure can be employed as a robust detection rule to identify adversarial examples. Accordingly, we develop a region growing algorithm to detect adversarial examples by checking whether there is an inconsistency in the current input.

\begin{figure*}[tb]
    \centering
    \includegraphics[width=0.85\textwidth]{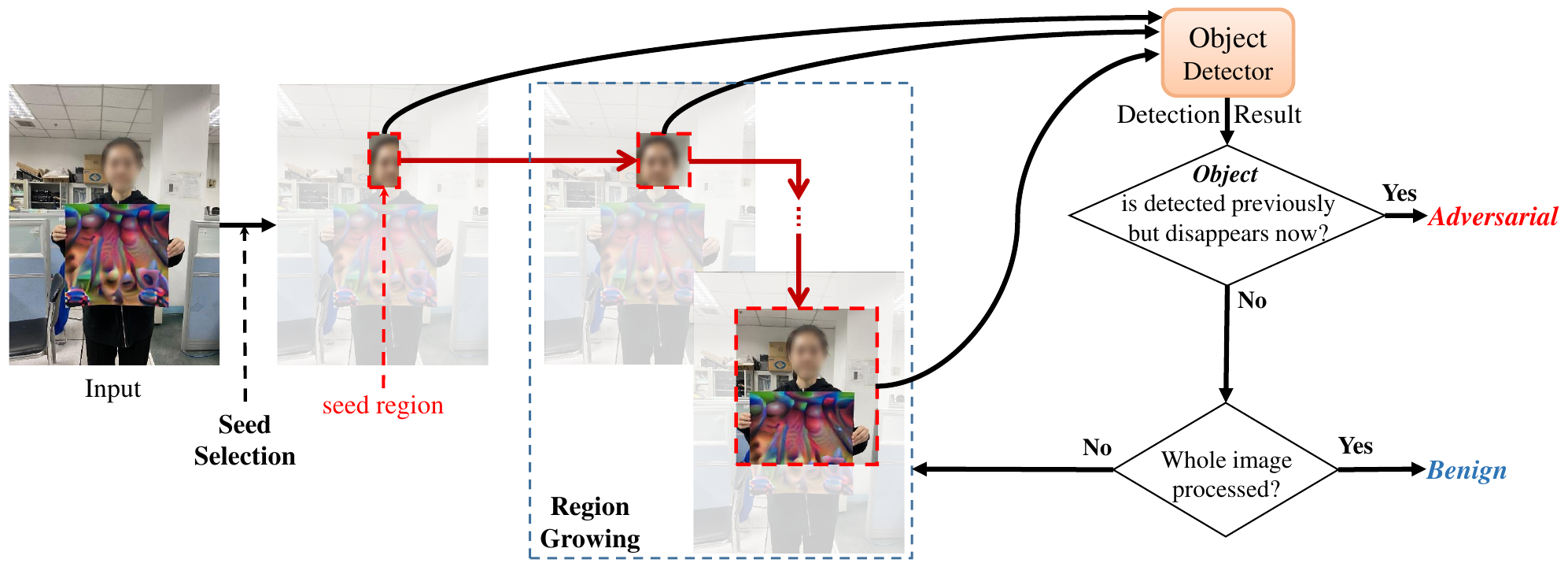}
    \caption{Detecting the adversarial example with the region growing algorithm.}
    \label{fig:growthframe}
\end{figure*}

More specifically, as shown in Figure~\ref{fig:growthframe}, the algorithm begins with a seed region, which is an image block selected from the input image. Iteratively, the region grows by including the neighbouring blocks following a proper growing direction. At each step, the grown region is fed to the object detector to check whether there is an object of interest (e.g., a person), and the detection result is recorded. When an object is detected in the previous region but missed in the current region, we believe the input is an adversarial example and the object identified in the last region will be output. In this way, the adversarial examples can be effectively detected without requiring any attack-specific knowledge.

\subsection{Content Semantics Consistency}
\label{subsec:inconsistency}

The state-of-the-art object detection models, e.g., YOLO and Faster R-CNN, are well trained with numerous samples and various data augmentation techniques, e.g., \textit{random cropping}, \textit{random erasing}, \textit{cutout} and \textit{gridmask}. This can make them able to successfully detect an object of interest even the input only contains a small part of the object. 
It is natural that when more parts are included in the input, the detection model is undoubtedly able to identify the object. 

\begin{figure}[!t]
    \centering
    \subfloat{
        \includegraphics[width=7cm]{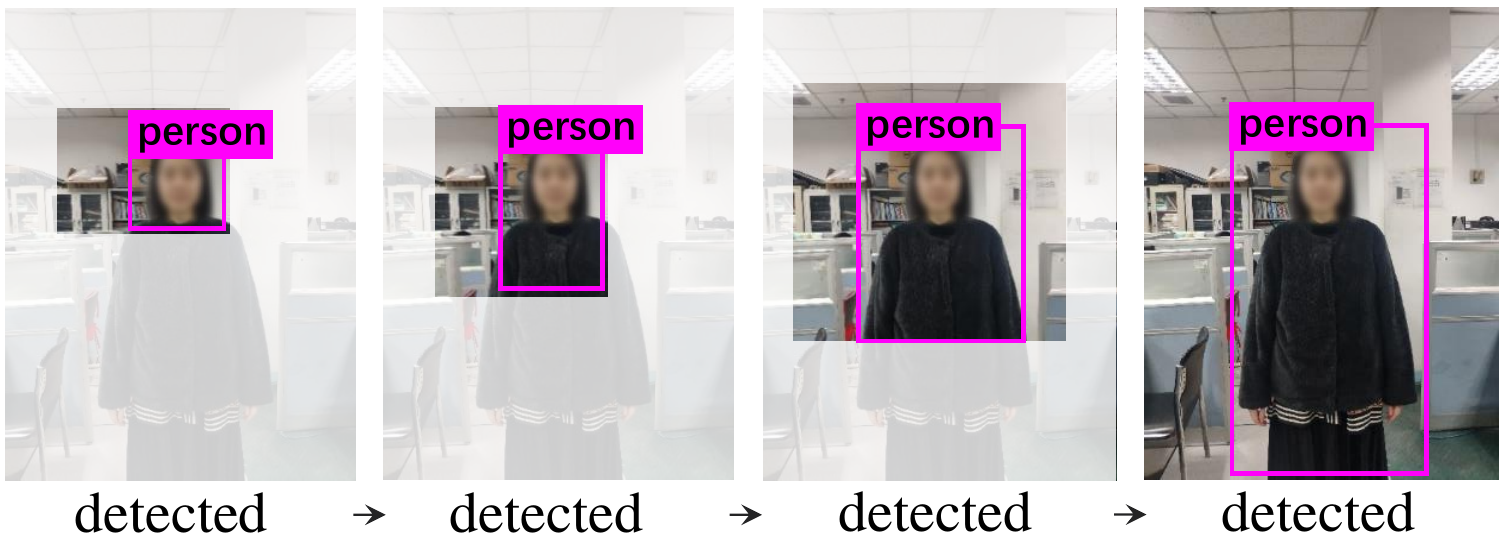}}
    \vspace{-3mm} \\ 
    \subfloat{
        \includegraphics[width=7cm]{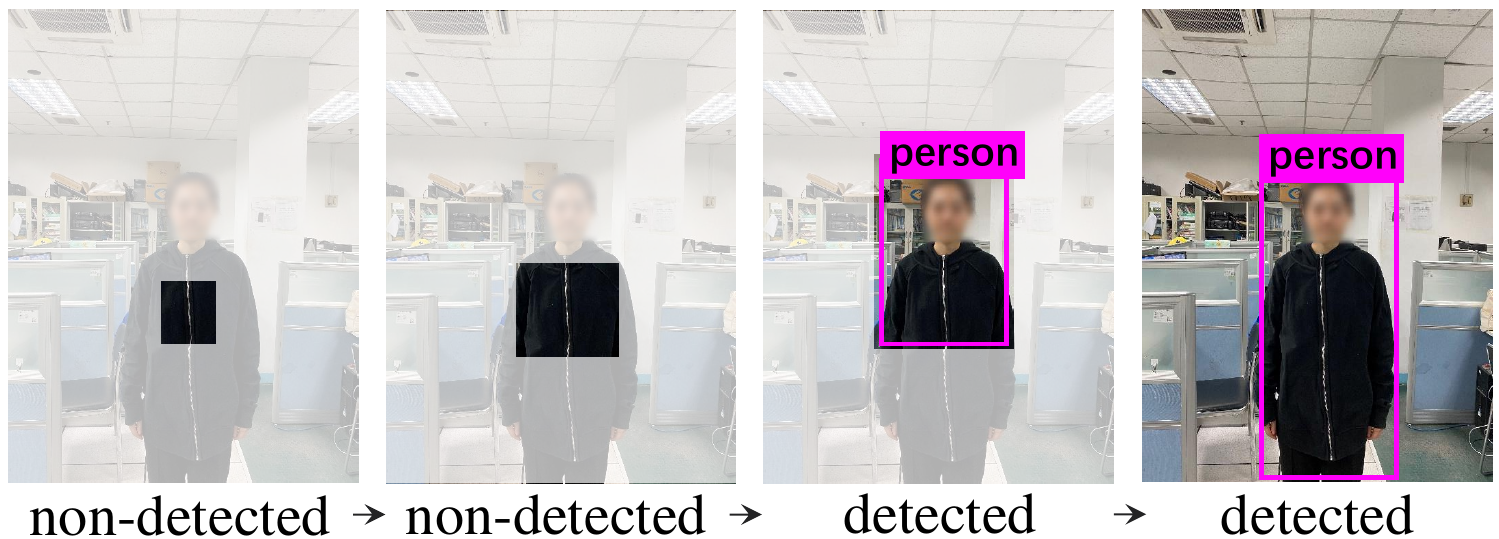}}
\caption{The content semantics consistency within the benign examples.}
\label{fig:ben_consistency}
\end{figure}

Take the person detection scenario as an example. If a person can be detected in a local part, the person can also be detected in global. We demonstrate it in Figure~\ref{fig:ben_consistency}, in which YOLO can consistently detect a person object from a small part, a larger part, and the whole of the input. Namely, \textit{if an object can be detected locally, it can also be detected globally}. We term the invariance between the local and global detection results \textit{the content semantics consistency}. The universal capability of modern detectors can be leveraged to distinguish adversarial patch examples from benign ones.

\begin{figure}[tb]
    \centering
    \includegraphics[width=7cm]{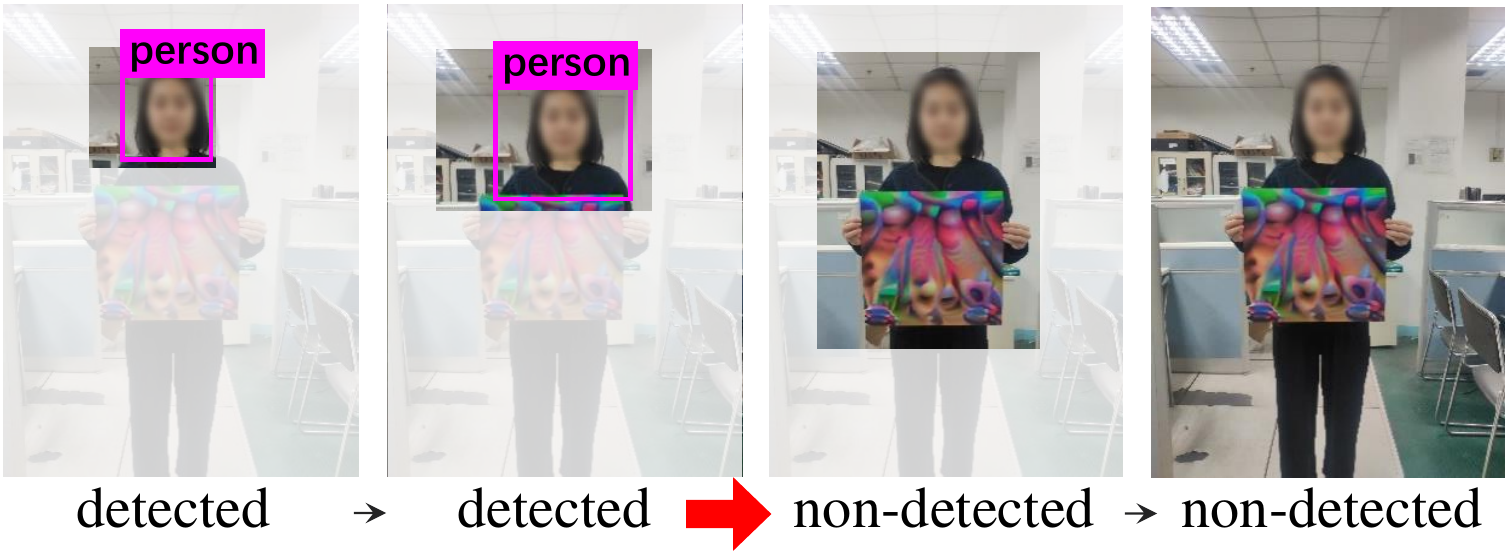}
    \caption{The content semantics inconsistency within the adversarial example.}
    \label{fig:adv_incon}
\end{figure}

For a benign example, the normal pattern of the detection result for the sequence of growing regions should be ``\textit{detected $\rightarrow$ detected $\rightarrow$ ... $\rightarrow$ detected}'' or ``\textit{non-detected $\rightarrow$ non-detected $\rightarrow$ non-detected $\rightarrow$ detected $\rightarrow$ detected $\rightarrow$ ... $\rightarrow$ detected}''. The former pattern indicates that the object can always be detected, and the latter indicates the object may be missed at first but is always hit starting from a certain region. Figure~\ref{fig:ben_consistency} shows the examples. The detected object should not suddenly vanish when given a larger input region. In other words, the normal detection state transition should be ``\textit{non-detected $\rightarrow$ non-detected}'', ``\textit{non-detected $\rightarrow$ detected}'', or ``\textit{detected $\rightarrow$ detected}'', but not ``\textit{detected $\Rightarrow$ non-detected}'' forever. 

However, this kind of content semantics consistency often does not hold in an adversarial example. As shown in Figure~\ref{fig:adv_incon}, for a person with an adversarial patch, she is more likely to be detected by an object detector just from a local part of her, but disappears when the grown region includes enough adversarial patch pixels. Consequently, a counter-common-sense detection state transition ``\textit{detected $\Rightarrow$ non-detected}'' emerges, which should never happen for a benign input.

On the basis of the above discussion, we are confident that the content semantics inconsistency can be leveraged as a detection criterion to effectively discover adversarial examples. We can monitor the trace of object detection results to check whether there is an abnormal detection state transition, i.e., ``\textit{detected $\Rightarrow$ non-detected}''.

It should be noted that the detection criterion is robust. 
In order to remove benign inconsistency, we perform further checks for inconsistent detections. Specifically, for each example, we first inspect the whole image, recording the position of all the detected "person" boxes. For each local image, when the box suspected of semantic inconsistency disappears, we compare and match the missing box with the detection box in the whole image recorded before. If there is no match, it is confirmed that there is inconsistency, indicating that the example is an adversarial example. If there is a match, it indicates that there is benign inconsistency in the local map, but not an adversarial example.
In practice, the attacker would not want to create an oversize adversarial patch for preserving its utility, giving the modern object detection model a chance to identify the adversarial object from its uncovered part. We will explain this detailedly in Section~\ref{subsec:4eva}.
More importantly, the detection criterion is general and independent of specific attack techniques. We can apply it to detect any kind of adversarial patches without caring how they are generated.

In addition to the content semantics consistency, there are also other types of semantics structures. Some of them can contribute to developing a high-performance detection algorithm. For example, we can directly count the human faces in the input and compare it with the count of detected person objects in the person detection task. If the two numbers are different, we can conclude that there is an adversarial attack. The face counting can be accomplished in real time with modern face recognition systems, e.g., ArcFace \cite{2019arcface}. Accordingly, the adversarial example detection can also be exceptionally efficient. Unfortunately, the approach is not robust enough and cannot be adopted in practice. The attackers can stick the adversarial patch on the back and turn their back to the camera. In this way, they can easily fool the person detection and avoid introducing abnormal face-person counts. We suggest that the robustness of a content semantics structure should be carefully investigated when employing it as a detection criterion.

\subsection{The Region Growing Algorithm}
\label{subsec:growth_strategy}

\begin{algorithm}[!t]\small
    \renewcommand{\algorithmicrequire}{\textbf{Input:}}
    \renewcommand{\algorithmicensure}{\textbf{Output:}}
    \caption{ \revbegin{1}{9/2-7/3-12} \nnewljc{Region Growing for Adversarial Patched Examples Detection} \revend{1}{9/2-7/3-12}}
    \label{alg:growth}
    \begin{algorithmic}[1]
        \REQUIRE $I$: An input image;  $Seed_g$: Growing seed. 
        \ENSURE $\varphi$: Indicating if an adversarial object exists.
        \STATE $Blocks = \textsc{ImagePartition}(I)$; $\varphi = False$
        \STATE $BProbabilities = \{\textsc{Model}_{Person}(b)~|~b \in Blocks\}$
        \STATE $BH2D = \{H_{2d}(b)~|~b \in Blocks\}$
        \STATE $Region_g = Seed$; $Dir = [up,~down,~left,~right]$ 
        \STATE $Blocks = Blocks-Seed$; $prob = \textsc{Model}_{Person}(Region_g)$
        \STATE $num = \textsc{Person}_{num}(Region_g)$
        \STATE \nnewljc{$num_{global} = \textsc{Person}_{num}(I)$}
        \WHILE{$\varphi == False$ \AND $Blocks \neq \emptyset$}
            \STATE $S = [0, 0, 0, 0]$
            \STATE $\forall d \in DIR, DBlocks[d] = \textsc{BlksInDir}(Region_g, d)$
            \IF{$prob < Person_{\textsc{Model}}$}
                \STATE $score_b = BProbabilities[b]$
                \STATE $\forall d \in Dir, S[d] = \sum_{b \in DBlocks[d]}{score_b}$
            \ELSE
                \STATE $score_b = BH2D[b]$
                \STATE $\forall d \in Dir, S[d] = \sum_{b \in DBlocks[d]}{score_b}$ 
            \ENDIF
            \STATE $d_g = arg\max_d{S[d]}$; $Region_s = Region_g$; $Region_n = \emptyset$
            \STATE $Score=BH2D$  if $num > 0$ else $BProbabilities$
            \STATE \nnewljc{$Region_n = \textsc{ARG}(Region_s, d_g, DBlocks[d_g], Score)$}
            \STATE $Blocks = Blocks - Region_n$; $Region_g = Region_n$
            \STATE $num = \textsc{Person}_{num}(Region_g)$
            \STATE \nnewljc{$\varphi = num > num_{global}  \big|_{Region_g}$}
        \ENDWHILE
    \end{algorithmic}
\end{algorithm}

 The detection mainly involves seed selection, growth direction determination, and region update. As described in Algorithm~\ref{alg:growth}, the region growing algorithm will terminate when an inconsistency is monitored or the region has grown to include all blocks.

\textbf{Seed selection.} The process begins by feeding an input image $I$ into the object detection model. \revbegin{3}{3/3-9} \nnewljc{The image $I$ is divided into $S \times S$ blocks (Line 1 in Algorithm~\ref{alg:growth}), where $S \times S$ represents the dimensions of the feature map produced as the output layer by the object detection model. For example, in YOLOv2, the image is mapped into 361 ($19\times 19$) blocks.} \revend{3}{3/3-9} Each block is associated with a classification probability output. Our experimental studies showed that randomly selecting a seed block often results in slow growth and can fail to effectively defend against attacks. Therefore, we adopt a heuristic strategy to select the growing seed. We choose the block with the highest \textit{Person} class probability as the seed $Seed_g$.

\textbf{Growth direction determination.} For $Seed_g$, the region $Region_{g}$ grows by incorporating additional blocks in a specific direction. This direction is determined based on the semantics of the blocks.

If no person object is detected in the current region $Region_{g}$, the region is expanded to include blocks with the highest \textit{Person} class probabilities. The sums of the probabilities from blocks in the four candidate directions—up, down, left, and right—are computed, and the direction with the highest total probability is selected as the growth direction $d_{g}$ (Lines 11 $\sim$ 12 in Algorithm~\ref{alg:growth}).

Conversely, if a person object is already detected in the current region, the goal is to rapidly encompass any potential adversarial patch to trigger an abnormal detection transition. The direction is determined using the image entropy $H_{2d}$ of the blocks. We hypothesize that an adversarial patch, to be effective, must contain significant "energy," implying high complexity and information density relative to normal input. Therefore, we use 2-D entropy \cite{abutaleb1989automatic} $H_{2d}$ to identify the most promising direction for detecting the adversarial patch (\nnewljc{Line 15} in Algorithm~\ref{alg:growth}).

For each pixel in an image block with 256 pixel levels (0 $\sim $ 255), the average pixel value of the 1-order neighborhood is first calculated. This forms a pair ($i$, $j$), the pixel value $i$ and the average of the neighborhood $j$. The frequency of the pair is denoted as $f_{ij}$, and a joint probability mass function $p_{ij}$ is calculated as
Equation~\ref{equ:2d-entr-p}, where $H$, $W$ is the height and the width of the block. On the basis, the 2-D entropy $H_{2d}$ of the block can be computed as Equation~\ref{equ:2d-entr-h}. Eventually, $H_{2d}$ of an RGB color block is the average of its three color planes' entropies, which are computed individually. In a similar way, the 2-D entropies of the blocks in a candidate direction are summed. The one with the highest $H_{2d}$ will be chosen as the next growth direction.

\begin{equation}
    \label{equ:2d-entr-p}
    p_{ij} = \frac{f_{ij}}{H \cdot W}, i \in [0, 255], j \in [0, 255]
\end{equation}
\begin{equation}
    \label{equ:2d-entr-h}
    H_{2d} = - \sum_{i=0}^{255} \sum_{j=0}^{255} p_{ij}\log_{2}(p_{ij}) 
\end{equation}

\begin{algorithm}[t]\small
    \renewcommand{\algorithmicrequire}{\textbf{Input:}}
    \renewcommand{\algorithmicensure}{\textbf{Output:}}
    \caption{\revbegin{1}{9/2-7/3-12}  \nnewljc{Adaptive Region Growth Algorithm (ARG)} \revend{1}{9/2-7/3-12}} \label{alg:getnewregion}
    \begin{algorithmic}[1]
        \REQUIRE $Region_s$: Current region; $d_g$: Growing direction; $DBlocks$: Candidate blocks in direction $d_g$; $Scores$: Score table for neighboring blocks. 
        \ENSURE $Region_g$: Updated region after growth. 
        \STATE $neighbor\_scores = \{Scores[b]~|~b \in DBlocks\}$
        \STATE $neighbor\_positions = \textsc{GetPositions}(DBlocks)$
        \STATE $l=\text{len}(neighbor\_scores)$
        \STATE $score_0 = \textsc{RegionScore}(Region_s, Scores)$  
        \STATE $pos_0 = \textsc{RegionPosition}(Region_s, Positions)$
        \STATE $g = [\,]$ 
        \FOR{$(score_i, pos_i) \in (neighbor\_scores, positions)$}
            \STATE $g_i = \dfrac{score_i - score_0}{pos_i - pos_0}$
            \STATE $g = g \cup \{g_i\}$ 
        \ENDFOR
        \STATE $G = sigmoid(\frac{1}{l} \sum g_i), g_i \in g$ 
        \STATE $growth\_step = \text{round} \left( (1 - G) \cdot k \right)$ 
        \STATE $Region_g = \textsc{UpdateRegion}(Region_s, DBlocks, growth\_step)$
        \vspace{-\baselineskip} 
        \RETURN $Region_g$
    \end{algorithmic}
\end{algorithm}

\textbf{Region update.} After determining the growth direction $d_{g}$, we propose the idea of \textit{Information Variation Rate} to compute how many blocks $growth\_step$ should be merged into the region $Region_{g}$. The \textit{Information Variation Rate} $G$ resulting from incorporating neighboring candidate blocks is calculated based on either the \textit{Person} class probability or $H_{2d}$. 
We have designed an adaptive Algorithm \ref{alg:getnewregion} for calculating growth step sizes. By measuring the \textit{Information Variation Rate} $G$ of the score in the growth direction, we determine the optimal step size. 
If $G$ is high, it indicates abundant information density in the current region, requiring slower growth (i.e., a smaller $growth\_step$). Conversely, a low $G$ suggests that the current $growth\_step$ is too large and should be adjusted downward. The maximum allowable step size $k$ for each growth iteration is defined as $k = round(0.25*s)$, meaning the step length in a single iteration cannot exceed 25\% of the image size (height or width) $s$. In Algorithm \ref{alg:getnewregion}, Lines 7 $\sim$ 11 show the calculation of $G$, and Line 12 shows how to get $growth\_step$ from $G$ and $k$. Here, $score_i$ is the \textit{Person} score or $H2D$ score of the block in position $pos_i$. 

\textbf{Attack detection.}
We first detect the global image and record all the detected object boxes as benign boxes, along with the total number of objects. During the region growing process, the abnormal detection state can be defined as the appearance of an anomalous object box in the local detection result that does not exist in the global image at the same position. In this case, the number of detected objects will decrease when comparing the local region's detection result to the global detection result in the same position. We consider it an inconsistency resulting from an attack (Lines 22 $\sim $ 23 in Algorithm~\ref{alg:growth}). 

To prevent false positives from benign inconsistencies, we use the global image to detect all detection boxes $Box$ before growing. During the growing process, if any detection box disappears but still exists in $Box$, it is treated as a benign inconsistency, and the growing continues. Otherwise, when an anomalous box is detected, it means there is a true inconsistency, triggering an alert for an attack. The threshold of Overlap Ratio $th_{or}=0.5$ is employed to verify if a disappeared box exists in $Box$.
Note that, the detection process handles individual or crowd scenes in the same say.

\subsection{Evaluation}
\label{subsec:4eva}

\subsubsection{Experiment settings} 
To evaluate the signature-independent defense against hiding person attacks, we conducted experiments using the same datasets in the digital and physical world, as those utilized in Section~\ref{subsec:3eva}, which comprise both adversarial and benign examples. 
Furthermore, to investigate the generality of the signature-independent method, we also try to detect another type of patch attack, which is designed to camouflage an object to another category rather than hiding it, e.g., dressing up a person as a dog \cite{huang2020UPC}.

The metrics measure the ability to detect whether an attack exists in a given sample. A detection is considered correct if an adversarial example is identified as adversarial~(\textit{True Positive}), or a benign example is identified as benign~(\textit{True Negative}). Otherwise, a \textit{False Negative} or \textit{False Positive} is reported. Then we calculate the precision (P), recall (R) and F1 score (F1).

\begin{table}[htbp]
\centering
\caption{\revbegin{2}{1} \nnewljc{The detection results of the signature-independent method.} \revend{2}{1}}
\label{table:4eva}
\begin{tabular}{c|c|c|ccc}
\hline
& Model & Dataset & P     & R      & \textbf{F1}     \\ \hline
\multirow{8}{*}{Digital} & \multirow{2}{*}{YOLOv2} & COCO & 0.943 & 0.881 & \textbf{0.911} \\ 
                   & & VOC  & 0.950 & 0.844 & \textbf{0.894} \\ 
& \multirow{2}{*}{YOLOv4} & COCO & 0.906 & 0.967 & \textbf{0.935} \\ 
                   & & VOC  & 0.962 & 0.862 & \textbf{0.909} \\ 
& \multirow{2}{*}{YOLOR} & COCO & 0.922 & 0.887 & \textbf{0.904} \\ 
                   & & VOC  & 0.925  & 0.949 & \textbf{0.937} \\ 
& \multirow{2}{*}{YOLOv8} & COCO & 0.944  & 0.919 & \textbf{0.931} \\ 
                  & & VOC  & 0.872  & 0.897 & \textbf{0.884} \\ \hline
\multirow{4}{*}{Physical} & \multirow{4}{*}{YOLOv2} & Indoor  & 1.000 & 0.877 & \textbf{0.934} \\ 
                   & & Outdoor  & 1.000 & 0.937 & \textbf{0.967} \\ 
                   & & \textcolor{black}{MyAdvTshirt}  & 0.986 & 0.944 & \textbf{0.965} \\
                   & & AdvTshirt \cite{xu2020T-shirt} & 1.000 & 1.000 & \textbf{1.000} \\ \hline
\multicolumn{3}{c|}{Total} & 0.951 & 0.904 & \textbf{0.926} \\ \hline
\end{tabular}
\end{table}

\subsubsection{Detecting hiding attack} 
As shown in Table~\ref{table:4eva}, the precision is always above 0.872, with 11 out of 12 cases higher than 0.9. Half of the cases (6 out of 12) have recalls higher than 0.9 and even the lowest recall of the others is as high as 0.844. As a consequence, the overall performance (F1) is never lower than 0.884.

As mentioned in Section \ref{sec:3sig}, we prepared adversarial clothes to generate adversarial examples, MyAdvTshirt. Signature-independent method detection results shows that our method is also effective on the detection of adversarial clothes, i.e. MyAdvTshirt and AdvTshirt, achieving F1-scores of 0.965 and 1.000 respectively. This experiment shows the robustness of our detection against clothing-based adversarial examples.

\begin{figure}[htbp]
    \centering
    \subfloat[Adversarial Example 1: an attacker and a benign object]{
        \includegraphics[width=0.95\linewidth]{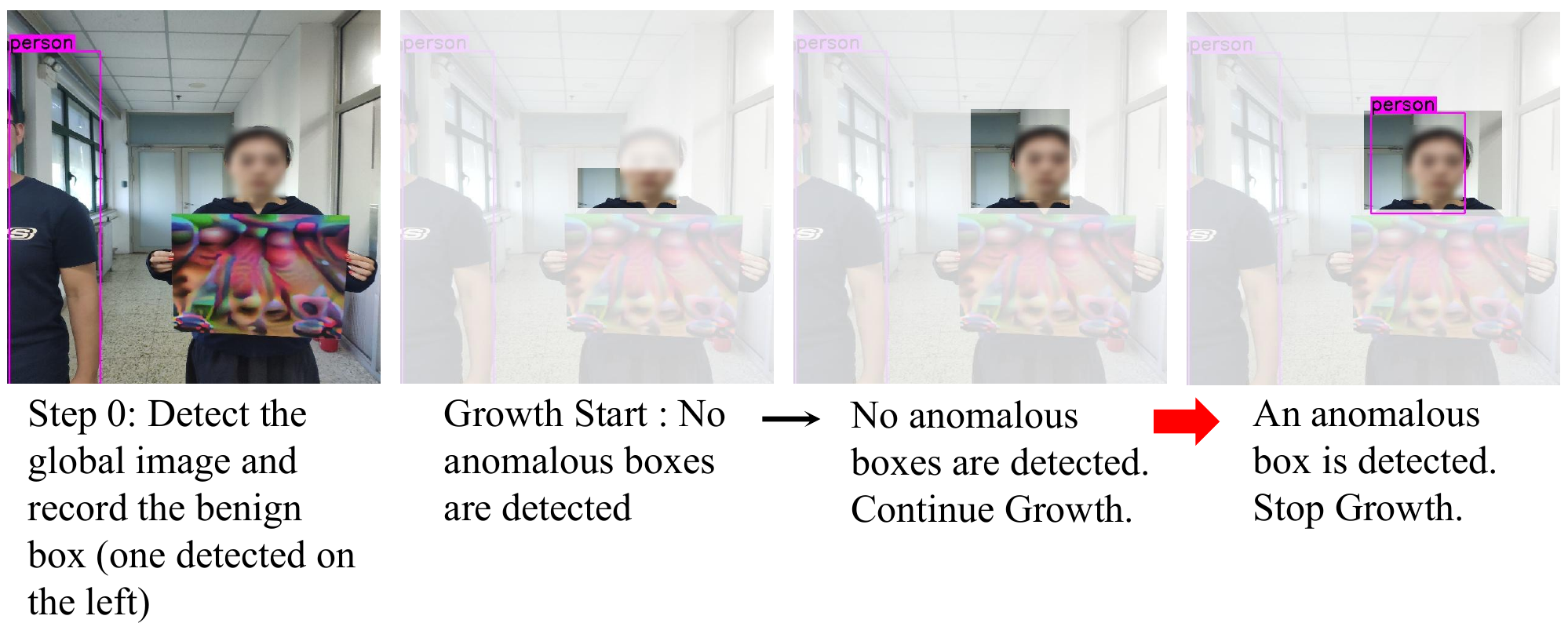}
        }
    \quad
    \subfloat[\revbegin{2}{1} \nnewljc{Adversarial Example 2: MyAdvTshirt} \revend{2}{1}]{
        \includegraphics[width=0.95\linewidth]{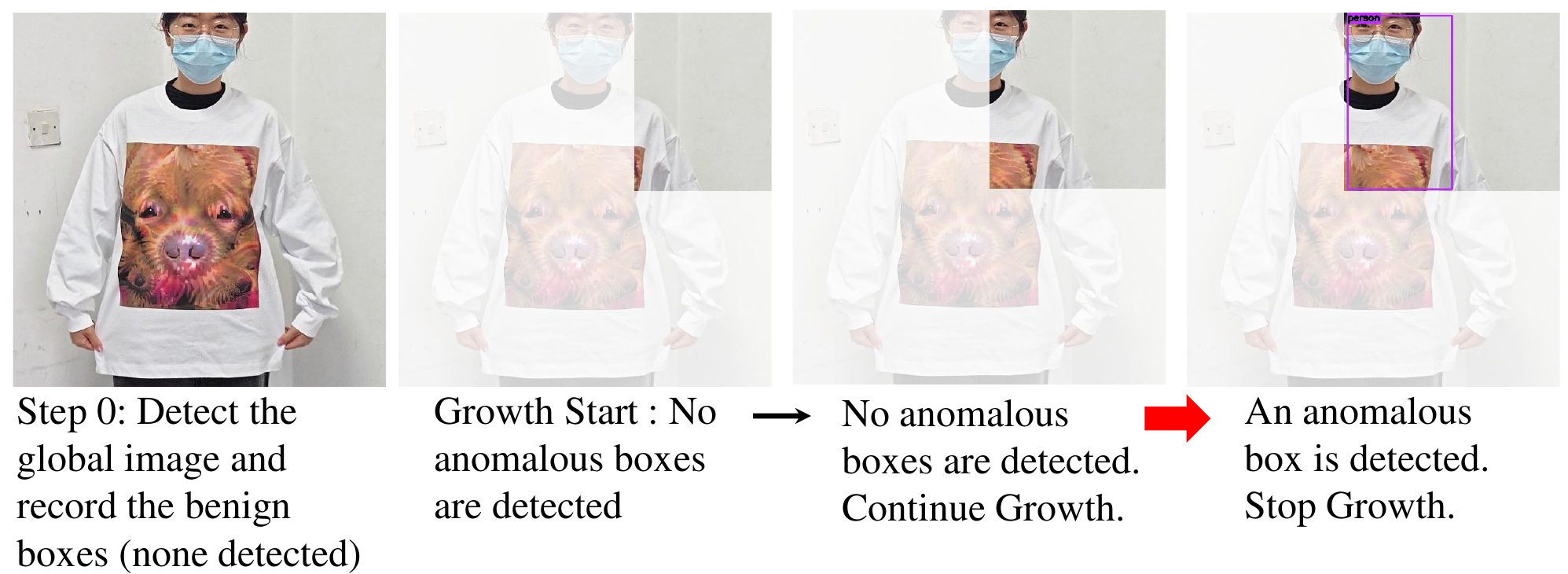}
        \label{fig:growth_myadvT}
        }
\caption{Inconsistency occurs in the physical adversarial examples.}
\label{fig:in_outdoor_detected}
\end{figure}

Figure~\ref{fig:in_outdoor_detected} illustrates two examples of adversarial sample detection in the physical world: Adversarial Example 1 from our physical world dataset using the AdvPatch attack, and Adversarial Example 2 from MyAdvTshirt. 

As to the comparisons, which are shown in Table~\ref{table:3 detection evaluation}, the overall performance (F1) of our signature-independent method outperforms DetectGuard, ObjectSeeker and UDF in 12, 11 and 12 cases respectively, out of all the 12 cases. 

\subsubsection{Detecting camouflage attacks}
\begin{figure}[tb]
    \centering
    \includegraphics[width=0.3\textwidth]{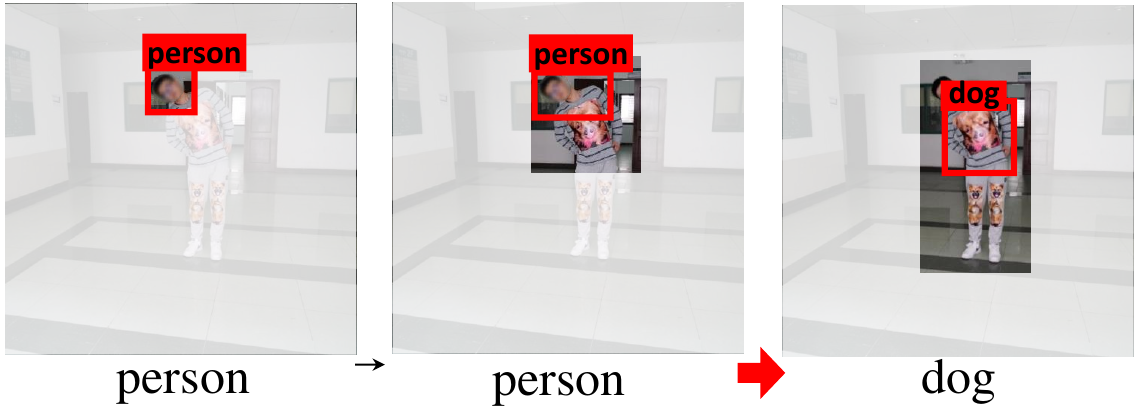}
    \caption{The signature-independent method detects the UPC adversarial example.}
    \label{fig:upc_growing}
\end{figure}

Huang et al. \cite{huang2020UPC} propose an attack method named Universal Physical Camouflage Attack (UPC). It can fool Faster R-CNN object detection model \cite{2017FasterRCNN} to misidentify an object as a wrong category with multiple camouflages rather than hiding it, such as detecting a person as a dog. Although our method is originally designed to defend the hiding object attacks such as AdvPatch, we find that our signature-independent method can be effectively applied to the camouflage attack and Faster R-CNN.
As shown in Figure~\ref{fig:upc_growing}, the camouflage example can also be successfully defended with the semantics inconsistency: the attack object is identified as a certain class locally but another one globally (\textit{a person locally but dogs globally}). 

We get a data set from UPC authors, which has 76 camouflage examples. We employed Faster R-CNN \cite{2017FasterRCNN} model with a detection threshold of 0.5 to evaluate the UPC examples. Our analysis revealed that only 40 out of the 76 adversarial examples successfully attacked the target detection model (causing misdetection), while the remaining 36 images failed to achieve successful attacks. Since our study focuses on detecting effective adversarial attacks, we only included the 40 successful adversarial examples in our experiments. 
We also randomly select 100 normal images from ImageNet \cite{deng2009imagenet} as the benign examples.
Table~\ref{tab:detectionUPC_growing} lists the results. 
With our method, 39 camouflage examples are effectively detected, and all benign examples are accurately identified. 

\begin{figure}[!t]
    \centering
    \includegraphics[width=0.5\textwidth]{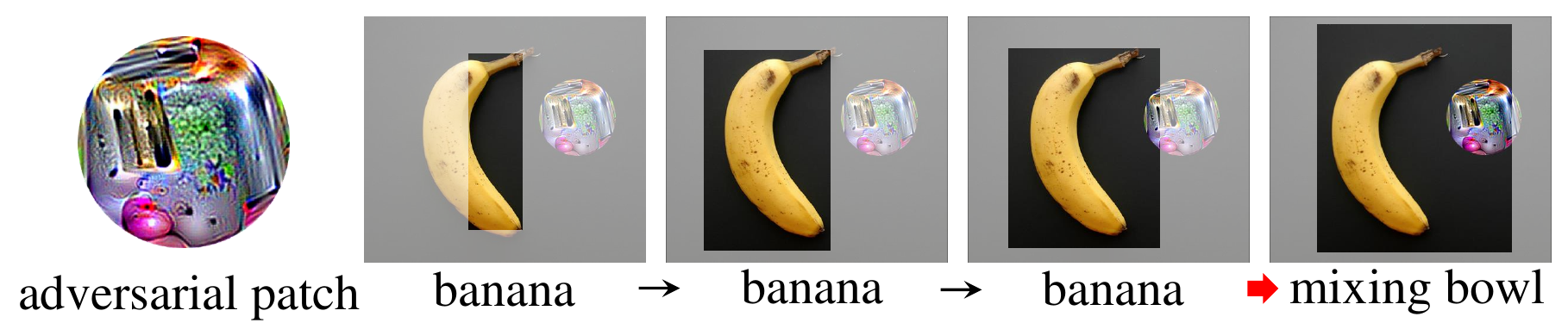}
    \caption{The content semantics inconsistency in the adversarial patched example targeting the image classification  \cite{brown2017advpatch}: \textit{a banana locally but a bowl globally}.}
\label{fig:adv_img_cls}
\end{figure}

Our idea can also be leveraged to counter other types of camouflage attack. We believe this study actually provides a general methodology for detecting local perturbation attacks. As another example, the adversarial patch targeting the image classification \cite{brown2017advpatch} can mislead the VGG16 \cite{2014VGG} classifier to ignore certain items (e.g., a banana) within the image, and report another class (e.g., a bowl). For defending the attack, we can determine the seed region and growing directions according to the given class of interest (e.g., the \textit{Banana} class), and check whether there is such a semantics inconsistency: \textit{a banana locally but another globally}, as shown in Figure~\ref{fig:adv_img_cls}.

Furthermore, it is worth pointing out in particular, the proposed method is not only signature-independent but also model-independent. It can be directly applied to detect attacks targeting different models except YOLO \cite{yolo9000}, i.e., Faster R-CNN \cite{2017FasterRCNN} and VGG16 \cite{2014VGG}.

\begin{table}[tb]
    \centering
    \caption{The detection results on UPC of the region growing algorithm.}
    \setlength{\tabcolsep}{4pt}
    \label{tab:detectionUPC_growing}
    \small
    \begin{tabular}{ c | c c c} \hline
         Data Set & Amount & Accuracy   \\ \hline
            Adversarial & 40 & 39 (97.5\%) \\
            Benign  & 100 & 100 (100\%)   \\ \hline
         Total & 140 & 139 (99.3\%) \\ \hline
    \end{tabular}
\end{table}

\subsubsection{Time cost analysis}
We conducted an analysis and statistical evaluation of the average detection time and average detection steps for adversarial examples and benign examples across various object detection models, as shown in Table \ref{tab: time cost sig-ind}. AEs and BEs represent adversarial examples and benign examples respectively. There is a major difference between adversarial examples and benign examples in terms of detection process and time consumption. The reason is that when detecting adversarial examples, the process stops growing once a local anomaly is detected, and the defense result is output immediately. However, when detecting benign examples, since there are no anomalies in the local image, the process continues to grow until it reaches the global image before stopping. Therefore, the number of detection steps for benign examples is greater than that for adversarial examples.

\begin{table}[!t]
    \centering
    \caption{\revbegin{1}{4/2-6} \nnewljc{Time cost of signature-independent method.} \revend{1}{4/2-6}}
    \label{tab: time cost sig-ind}
    \begin{tabular}{c|cccc}
    \hline
        Model &  AEs time$(s)$  & AEs steps & BEs time$(s)$   & BEs steps \\ \hline
        YOLOv2  & 0.469 & 5 & 1.454 & 15 \\ \hline
        YOLOv4 & 0.455 & 8 & 0.585 & 13 \\ \hline
        YOLOR & 1.005 & 6 & 1.996 & 17 \\ \hline
        YOLOv8 & 0.452 & 12 & 0.617 & 19 \\ \hline
    \end{tabular}
\end{table}

In addition, we analyzed the detection process in single and crowd scenes and their impacts on time consumption. We have evaluated the detection time for adversarial examples across three different scenarios: (a) single scenes; (b) crowd scenes with only one adversarial patch; and (c) crowd scenes with more than one patch. The average detection time and the number of steps were the smallest in situation (a). This is because the information density in such scenes is relatively low, resulting in larger adaptive step sizes, which in turn reduced the total detection time and the number of steps required. In contrast, situation (b) exhibited the largest average detection time and number of steps. Higher information density in crowd scenes leads to smaller adaptive step sizes and thus a larger total number of steps to detect anomalies. For situation (c), the average detection time and the number of steps were slightly lower than in situation (b). This can be attributed to the presence of multiple patches in these scenes, in which the probability of encountering an anomaly during the growth process is higher. Once an anomaly is detected, the detection process stops, resulting in shorter detection time. 

\subsubsection{Detection on defense-aware attack}

\begin{figure}[tb]
    \centering
    \includegraphics[width=0.22\textwidth]{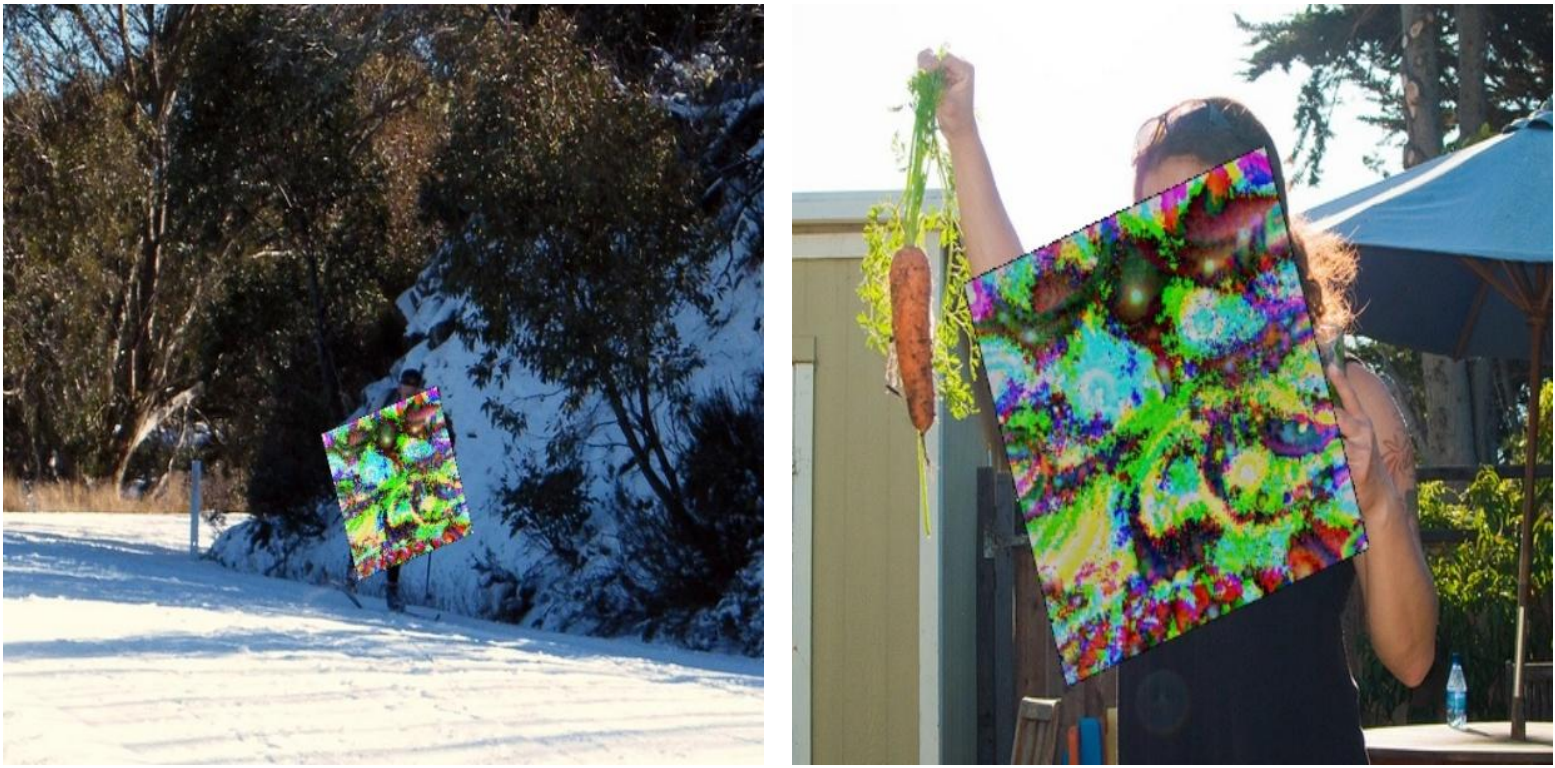}
    \caption{Failed cases with large-size patches.}
    \label{fig:growth_largesizepatch}
\end{figure}

We attempted a defense-aware attack on the patch's original size, however, we were unable to successfully attack the Region Growing detection under any circumstances. Therefore, we tried to increase the patch size so that region detection would be affected by patch in each step of the local growth map. In the original adversarial sample, the size parameter of generated patch is $size_n=0.2$. We generated three kinds of enlarged patches: large patch, larger patch, and largest patch, with $size_n$ of 0.4, 0.6, and 0.8 respectively. Our experiments prove that our method still has some defense capability in the case of large and larger patches, although under these samples, the patches are already large to the point of being unreasonable.
When $size_n=0.4$, the large patch size is almost twice the original patch size, and the detection accuracy is 76.1\%. When $size_n=0.6$, the larger patch size is almost three times the original patch, and the detection accuracy is 44.9\%. When $size_n=0.8$, the largest patch size is almost four times the original patch, and the detection accuracy drops to 22.0\%. However, as shown in Figure~\ref{fig:growth_largesizepatch}, the largest patch can successfully bypass the Region Growing detection while the person is almost completely blocked, and the utility of the sample is seriously affected, violating the principle that the attack should be as inconspicuous as possible. 

\subsubsection{Causes of FPs and FNs}
\begin{figure}[t]
    \centering
    \includegraphics[width=\linewidth]{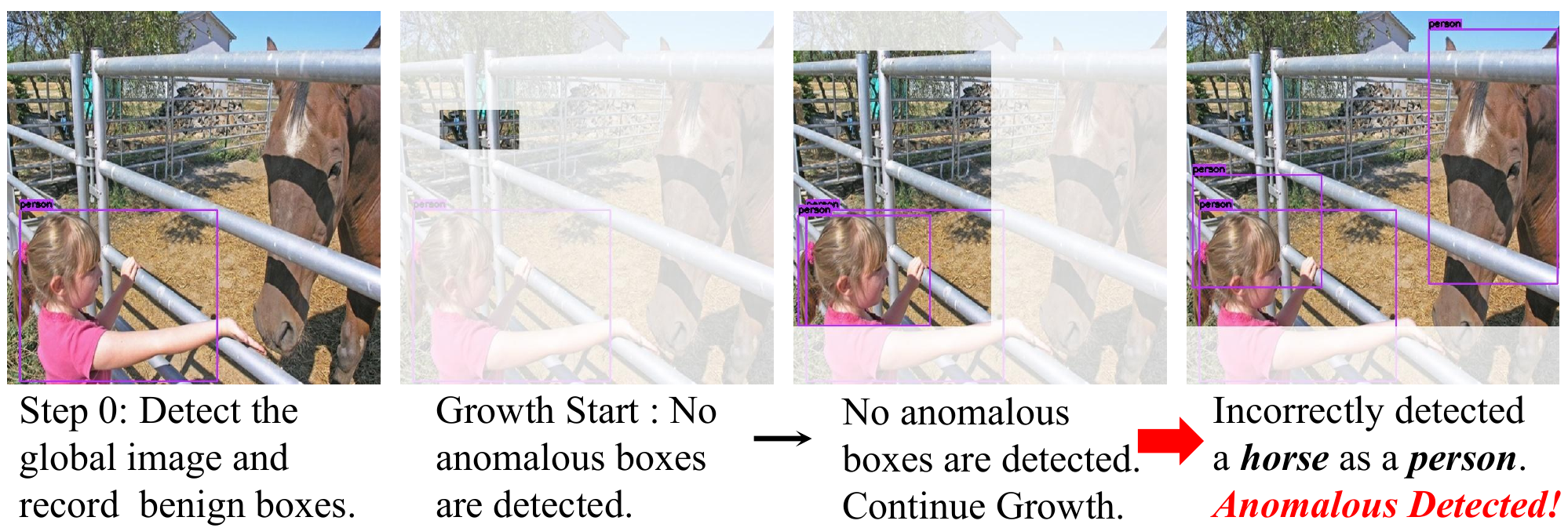}
    \caption{\revbegin{1}{2} \nnewljc{An FP of the signature-independent method.} \revend{1}{2}}
    \label{fig:sig_ind_FP}
\end{figure}

\begin{figure}[htbp]
    \centering
    \includegraphics[width=\linewidth]{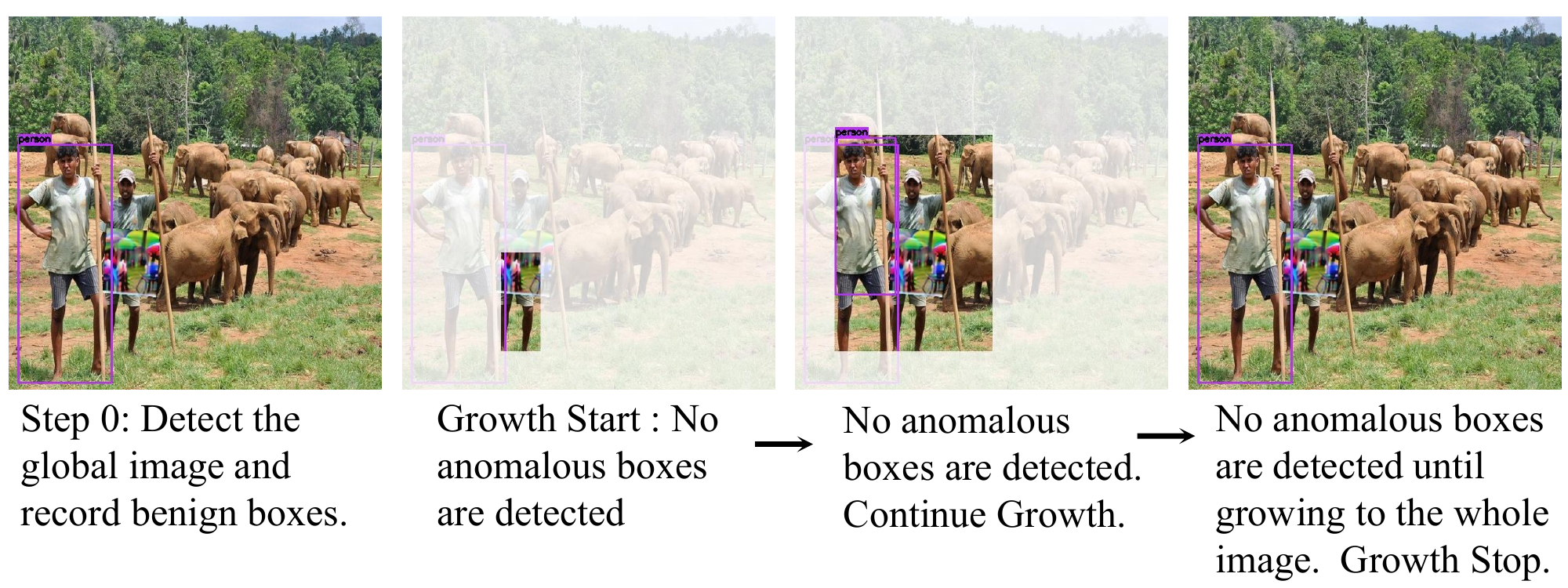}
    \caption{\revbegin{1}{2} \textcolor{black}{An FN of the signature-independent method.} \revend{1}{2}}
    \label{fig:sig_ind_FN}
\end{figure}

The main cause of FPs is that when detecting local images, even without any adversarial information, the detection model itself may sometimes incorrectly identify objects that do not actually exist, thus wrongly detected the anomalous box. An example is shown in Figure ~\ref{fig:sig_ind_FP}.
FNs are caused, in certain examples, because the target "person" is relatively small or lacks prominent features. For example, in Figure~\ref{fig:sig_ind_FN}, at the start of the growth process, the growth seed fails to locate the potential target and instead begins on the adversarial patch. Consequently, throughout the subsequent growth process, the adversarial patch remains within the localized image region, preventing the local semantics of the actual target from being properly detected. 

\subsubsection{Applicability to other object detection models}
To verify the generalizability of our signature-independent approach, we conducted extended experiments on the two-stage detector, i.e., Faster RCNN, and the transformer-based architecture, i.e., DINO~\cite{dino}. Table \ref{tab:sig_ind_other_models} shows the precision, recall and F1 score of our experiments on these models.

Using the adversarial examples and benign examples from the dataset of Feng et al. \cite{feng2025fire}, we evaluated Faster RCNN and found that its F1 score reached 0.909 on the digital dataset and 0.960 on the physical dataset, which consistent performance as the YOLO models.

Additionally, we performed experiments on DINO using the adversarial examples we generated from the COCO and VOC datasets, along with their corresponding benign examples.  Although the recall of DINO is slightly lower than YOLO models, achieving 0.792, which possibly dues to DINO occasionally neglecting local object features, its F1 score consistently exceeds 0.85, demonstrating the effectiveness of global shape analysis in general detection models.

\begin{table}[!ht]
    \centering
    \caption{\revbegin{3}{2/3-10} \textcolor{black}{The detection results on other models.}\revend{3}{2/3-10}}
    \label{tab:sig_ind_other_models}
    \begin{tabular}{c|c|ccc}
    \hline
        Model & Dataset & P & R & F1 \\ \hline
        \multirow{2}{*}{Faster RCNN} & Digital & 0.921 &  0.897  & 0.909  \\
         ~ & Physical & 0.985 & 0.936 & 0.960  \\ \hline
        DINO & Digital & 0.933 & 0.792 & 0.857 \\ \hline
    \end{tabular}
\end{table}

\subsubsection{Detection on other object classes}
We specifically selected \textit{Plane} and \textit{Cat} for additional validation. We apply YOLOv8 to COCO-based adversarial and benign examples, consistent with those described in Section \ref{subsub:other obj cls}. \textit{Plane} objects achieved P=0.968, R=0.882, F1=0.923, and \textit{Cat} objects achieved P=0.938, R=0.968, F1=0.952. These results show that our method is also effective on other object types.

\section{Discussion}
\label{sec:discussion}

\subsection{Application scenarios of the two methods}
We proposed two independent methods based on their distinct characteristics and application scenarios. The signature-based method offers rapid detection that can detect 6-7 frames per second. This makes it suitable for scenarios requiring real-time detection. However, it should be noted that the signature-based method requires prior knowledge of attack features and is therefore limited to defending against known adversarial patch attacks. In contrast, the signature-independent method operates at a relatively slower speed. This method is more appropriate for scenarios where real-time detection requirements are less stringent. The significant advantage of the signature-independent method is that it does not require knowledge of attack signatures during the detection process. Consequently, it is effective not only against known adversarial patch attacks but also capable of detecting novel, previously unseen adversarial patch attacks.

\subsection{Identifying signatures} 
As mentioned above, the key of enforcing the signature-based defense is to find discriminating signatures. An important problem is how to identify the leverageable signature as soon as possible when facing unknown attacks. Using the differential analysis to investigate the benign and adversarial example may be a feasible way, and the visualization technique can be used as an observation tool to discover the interesting difference easily.  Note that the emerging hiding attacks and camouflage attacks can be caught by using the signature-independent detection.

\subsection{Diversity of the patches} 
To demonstrate the diversity of our attack scenarios, we employed 10 different adversarial patches in our experiments. These patches were designed to target various models, employ various attack methods, and focus on different object categories. Both placard patches and clothing patches were included. Our experiment results show that our defense method effectively detects adversarial samples across all the diverse adversarial patches tested, confirming the robustness of our approach regardless of the specific patch design or attack methodology.

\subsection{Time overhead}  
In general, the signature-independent method is relatively slower than the signature-based method, i.e., about 0.96s of YOLOv2 and 0.53s of YOLOv8 per image on average. The detailed time cost is analyzed in Table~\ref{tab: time cost sig-ind}.

To further enhance detection efficiency, we also implemented engineering optimizations to improve the detection speed by employing the parallel computing. Table~\ref{tab: parallel time cost sig-ind} shows the average detection time and average steps for each model after applying parallel computing. Here, AEs represent adversarial examples, and BEs represent benign examples. 
\begin{table}[!t]
    \centering
    \caption{Time cost of signature-independent method after parallel computing.}
    \label{tab: parallel time cost sig-ind}
    \begin{tabular}{c|cccc}
    \hline
        Model &  AEs time$(s)$  & AEs steps & BEs time$(s)$   & BEs steps \\ \hline
        YOLOv2  & 0.342 & 5 & 1.183 & 15 \\ \hline
        YOLOv4 & 0.422 & 8 & 0.565 & 13 \\ \hline
        YOLOR & 0.565 & 6 & 1.500 & 17 \\ \hline
        YOLOv8 & 0.419 & 12 & 0.580 & 19 \\ \hline
    \end{tabular}
\end{table}
Although it is relatively slower, it is also acceptable in some noncritical cases. We can limit the number of examples needed to be detected via sampling the input frames, especially when the signature-independent method is used to catch the unknown attack examples. We can design an aggressive growing strategy to obtain multiple candidate regions simultaneously. Besides, combining the two methods to get a more comprehensive defense is also reasonable. This is one of our future works.

\subsection{Negative correlation between confidence scores and object scales} 
In object detection, the object confidence score produced by the detector is also valuable for identifying adversarial patches. 
We have validated the idea that the negative correlation between object confidence and the scale of detected objects suggests the presence of adversarial patches. We performed the experiment on YOLOv2, using 168 adversarial examples generated from COCO dataset. Direct implementation of size-confidence correlation monitoring yielded limited success: achieved recall rate of approximately 0.1 and suffered from significant false positives. While this preliminary approach shows insufficient detection capability, its core idea is very insightful and it can be integrated into the adaptive region growth of our signature-independent method. Specifically, the observed negative correlation between confidence scores and object sizes will guide adaptive adjustments in our future work.

\subsection{Guided Grad-CAM based on other output classes}

The Guided Grad-CAM method is primarily used for detecting known attacks, which are somewhat related to the specific content of adversarial patches. In practice, we can enhance performance by selecting output classes that align with the information distribution of adversarial patches specific to different models when calculating the Guided Grad-CAM score. We observed the information distribution of adversarial patches on YOLOR and YOLOv8, and incorporated the \textit{Donut} output class into the calculation of the Guided Grad-CAM score. As shown in Table \ref{fig:ggc_on_other_cls}, this simple adjustment significantly improved the recall of the method. We believe that adopting a more attack-specific score calculation approach could further enhance the performance, and we plan to explore this direction in future research. 

\begin{table}[!ht]
    \centering
    \caption{Guided Grad-CAM-based detection results based on other output classes.}
    \label{fig:ggc_on_other_cls}
    \begin{tabular}{ccccc}
    \hline
        Model & Dataset & P & R & F1 \\ \hline
        \multirow{2}{*}{YOLOR} & COCO & 0.909 & 0.690 ( \textcolor{red}{$\uparrow $ 0.332} ) & 0.784 \\
       ~ & VOC & 0.889 & 0.615 ( \textcolor{red}{$\uparrow $ 0.051} ) & 0.727 \\ 
        \multirow{2}{*}{YOLOv8} & COCO & 0.952 & 0.633 ( \textcolor{red}{$\uparrow $ 0.226} ) & 0.761 \\ 
        ~ & VOC & 0.973 & 0.572 ( \textcolor{red}{$\uparrow $ 0.188} ) & 0.720 \\ \hline
    \end{tabular}
\end{table}

\section{Conclusion}

To defend the adversarial patch attacks aiming at object detection, in this paper, we present two different defense methods, one is signature-based and the other is signature-independent. At first, we find that many adversarial patch pixels possess a discriminating characteristic. Accordingly, a defense filter is developed to exclude these patch pixels from the input such that the filtered adversarial patch loses its attack efficacy. This signature-based defense is demonstrated to be effective and time-efficient for the existing attacks. However, it might be ineffective against the potential risks. To provide a robust defense against potential and unknown attacks, we propose a general signature-independent detection method. The method is based on an insightful observation that the global semantics of an adversarial patched example is inconsistent with its local semantics. The signature-independent detection has been proven to be effective against known and unknown attacks, although it is relatively slow.
Essentially, the two defenses are complementary to each other and can be used for different scenarios, or be combined to get a comprehensive protection.



\ifCLASSOPTIONcaptionsoff
  \newpage
\fi

\bibliographystyle{abbrv}
\bibliography{ref}

\begin{IEEEbiography}[{\includegraphics[width=1in,height=1.25in,clip,keepaspectratio]{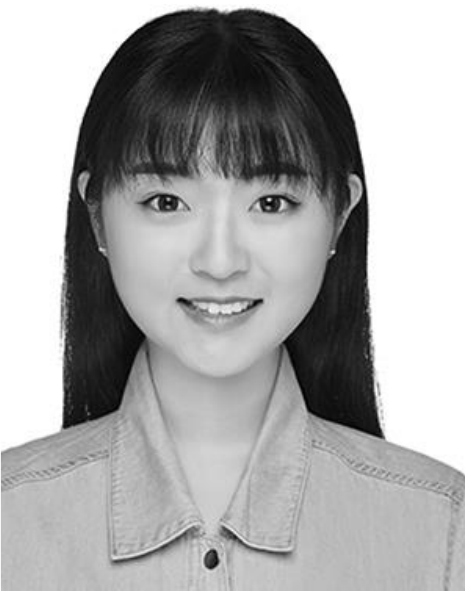}}]{Jiachun Li} 
received the BS degree in communication engineering from the University of Science and Technology Beijing, Beijing, China. She is currently working toward the Ph.D. degree in information security with the School of Information, Renmin University of China, Beijing, China. Her research interests focus on adversarial machine learning and AI security.
\end{IEEEbiography}

\begin{IEEEbiography}[{\includegraphics[width=1in,height=1.25in,clip,keepaspectratio]{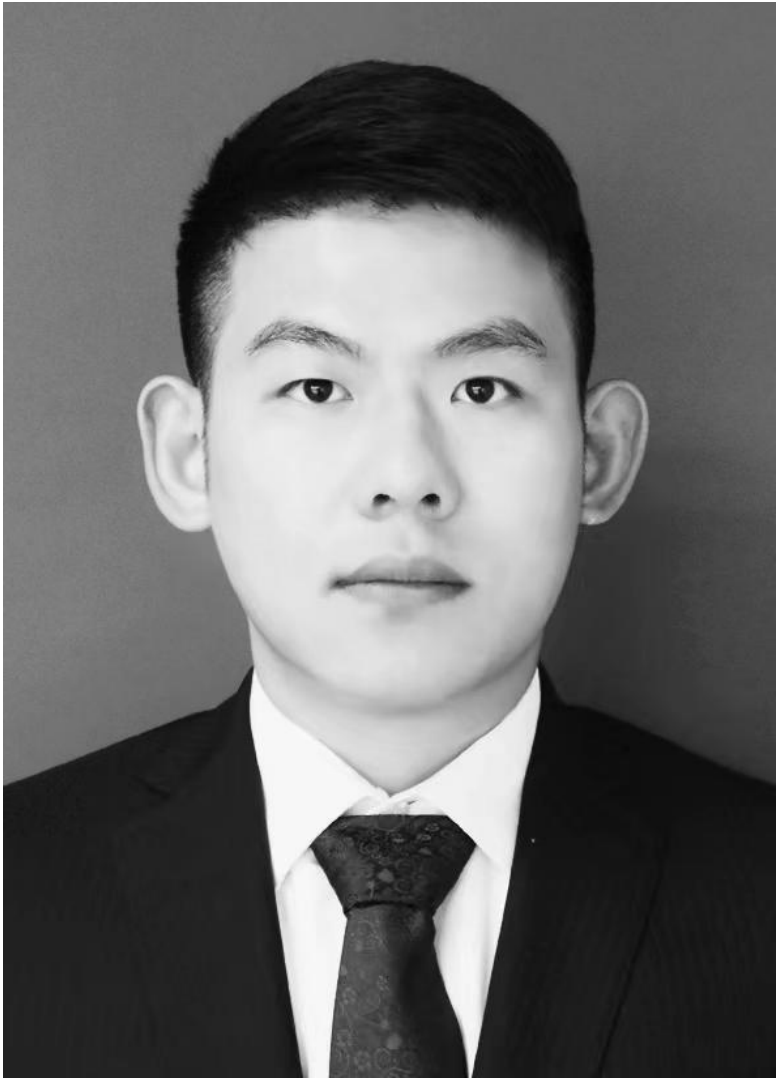}}]{Jianan Feng} 
 received his Master Degree in Software Engineering from Yunnan University. He is currently working toward the Ph.D. degree in information security with the School of Information, Renmin University of China, Beijing, China. His research interests focus on adversarial machine learning and AI security.
\end{IEEEbiography}

\begin{IEEEbiography}[{\includegraphics[width=1in,height=1.25in,clip,keepaspectratio]{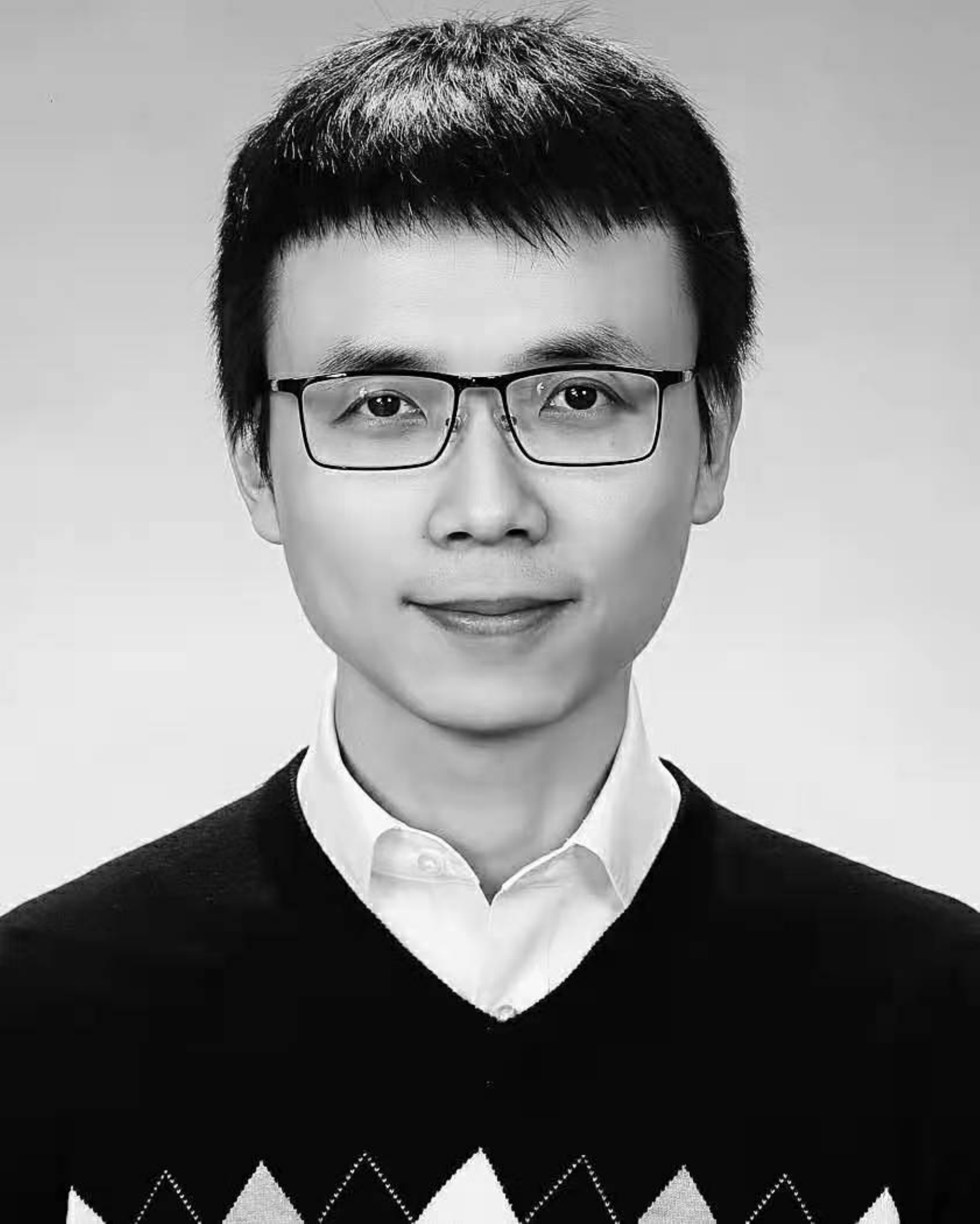}}]{Jianjun Huang} 
  received the Ph.D. degree in Computer Science from Purdue University. 
  He is currently an associate professor at School of Information, Renmin University of  China. His research interests focus on program analysis, vulnerability detection, 
  mobile security and blockchain security. 
\end{IEEEbiography}

\begin{IEEEbiography}[{\includegraphics[width=1in,height=1.25in,clip,keepaspectratio]{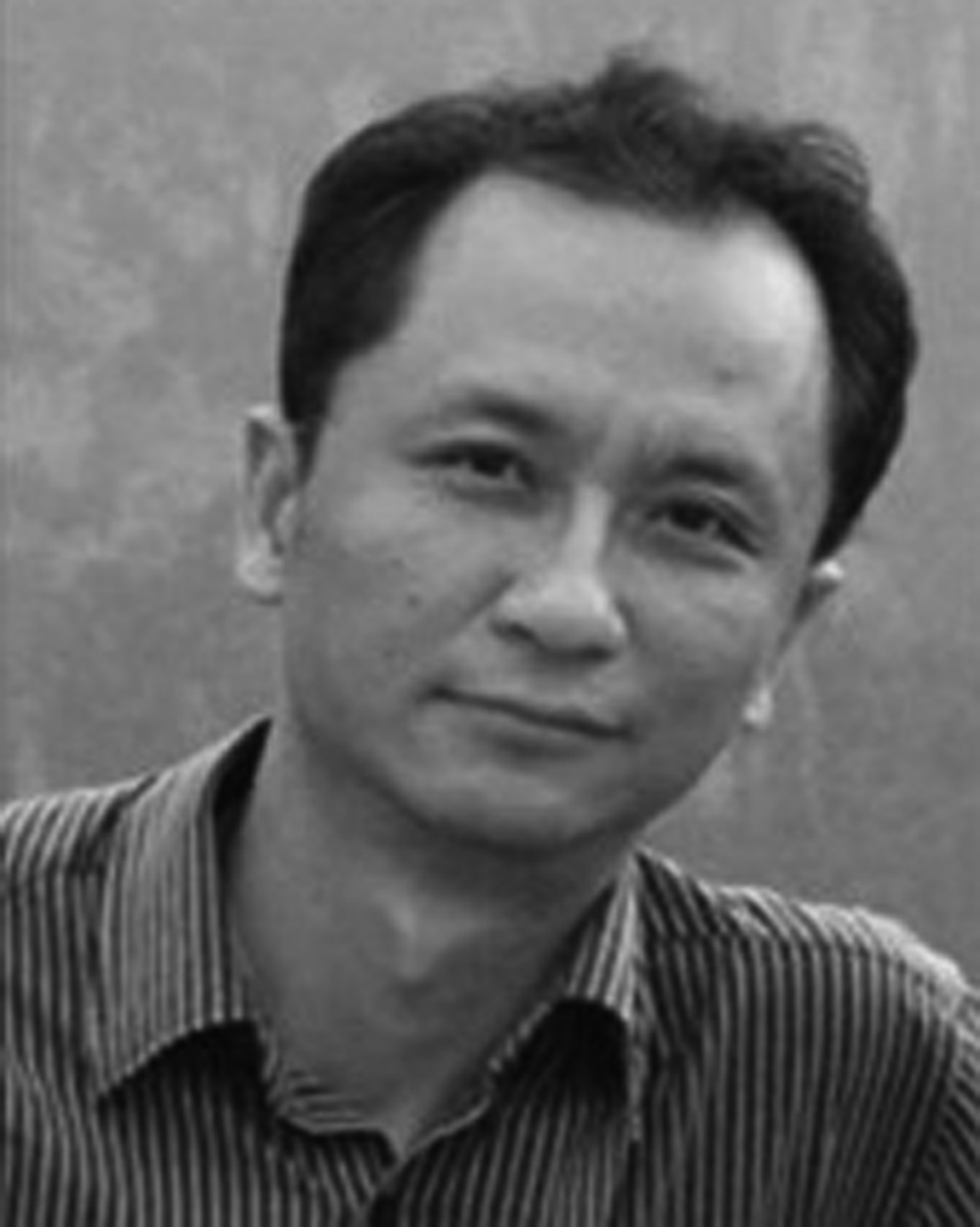}}]{Bin Liang} 
  received the Ph.D. degree in Computer Science from Institute of Software, 
  Chinese Academy of Sciences. He is currently a professor at School of Information, 
  Renmin University of China. His research interests focus on program analysis, 
  vulnerability detection, mobile security, and AI security.
\end{IEEEbiography}

\end{document}